%% file: main.tex
\documentclass[a4paper,11pt,twoside]{article}

\usepackage{amsmath}
\usepackage{hyperref}
\hypersetup{colorlinks,linkcolor={blue},citecolor={red},urlcolor={blue}}
\usepackage{xcolor}

\usepackage[a4paper,hscale=0.7,vscale=0.75,centering]{geometry}
\usepackage{fullpage}
\usepackage{authblk}
\usepackage[utf8]{inputenc}
\usepackage[T1]{fontenc}
\usepackage[english]{babel}
\usepackage{charter}

\usepackage{times}

\usepackage[utf8]{inputenc} 
\usepackage[T1]{fontenc}    
\usepackage{hyperref}       
\usepackage{url}            
\usepackage{booktabs}       
\usepackage{amsfonts}       
\usepackage{nicefrac}       
\usepackage{microtype}      
\usepackage{lipsum}
\usepackage[noabbrev, capitalise]{cleveref}
\usepackage{adjustbox}
\usepackage{cancel}

\usepackage[round]{natbib}

\usepackage{graphicx}
\usepackage{subfigure}
\usepackage{color}
\usepackage{relsize}
\input{abbrev.tex}

\title{Non-linear Multitask Learning with Deep Gaussian Processes}
\author[1,2]{Ayman Boustati}
\author[1,2]{Theodoros Damoulas}
\author[2]{Richard S. Savage}

\affil[1]{The Alan Turing Institute, London}
\affil[2]{University of Warwick}
\date{}

\begin{document}
\selectlanguage{english}
\maketitle

\begin{abstract}
 We present a multi-task learning formulation for Deep Gaussian processes (DGPs), through \emph{non-linear} mixtures of latent processes. The latent space is composed of \textit{private} processes that capture within-task information and \textit{shared} processes that capture across-task dependencies. We propose two different methods for segmenting the latent space: through hard coding shared and task-specific processes or through soft sharing with Automatic Relevance Determination kernels. We show that our formulation is able to improve the learning performance and transfer information between the tasks, outperforming other probabilistic multi-task learning models across real-world and benchmarking settings.
 \end{abstract}

\input{sections/introduction.tex}

\input{sections/model_specification.tex}

\input{sections/related_work.tex}

\section{EXPERIMENTAL RESULTS}
We test the three variations of the presented model (described in \cref{sec:model_instantiation}), which we collectively refer to as \textbf{MDGP}, on three datasets. For all of the experiments, we compare our proposal with three other GP-based models: \textbf{\stldgp{}}: independent single-task 2-layer DGPs \citep{damianou_deep_2013,salimbeni_doubly_2017} trained on each task individually; \textbf{\stlgp{}}: a independent single-task variational sparse Gaussian process  \citep{hensman_gaussian_2013} trained on each task individually; \textbf{\icmgp{}}: a multi-task variational sparse Gaussian process \citep{nguyen_collaborative_2014, bonilla_multi-task_2007}. This model uses the intrinsic coregionalisation kernel (ICM) \citep{goovaerts_geostatistics_1997} for multi-task learning.

Furthermore, we also introduce further comparisons to standard multi-task neural network of similar complexity \citep{bakker_task_2003}: \textbf{\mtlannTwo{}} and \textbf{\mtlannThree{}}: two and three layer multi-head feed-forward networks respectively, with uncertainty computed by MC Dropout \citep{gal2016dropout}; \textbf{\mtlbnnTwo{}} and \textbf{\mtlbnnThree{}}: two and three layer multi-head variational Bayesian neural networks respectively.

For all the models, we use the Mat\'ern-5/2 kernel and use the same number of inducing points per task for all the models. We jointly learn the variational parameters and hyperparameters (of the kernels -including ARD weights- and likelihoods) through the maximisation of the models' ELBO. We use the Adam Optimiser \citep{kingma_adam:_2015} on the DGP-based models, and L-BFGS \citep{nocedal_numerical_2000} for the shallow GP-based models (except when mini-batching was used where we revert back to Adam). See \cref{sec:experiment-details} for a more detailed description of the experimetal setup. 

\input{sections/mnist_variations.tex}
\input{sections/sarcos.tex}
\input{sections/faims.tex}

\section{CONCLUSION}
In this work, We introduced a new framework for multi-task learning for DGPs, presenting a general formulation of a multi-process layer structure that can learn shared information between the tasks, as well as task specific information. Inference in this model is possible through the multi-task extension to the doubly stochastic variational approximation, with practically fast inference in CPU time (\cref{table:timing} in \cref{sec:appendix_sarcos}). We proposed three instantiations of this framework with different properties. The experimental results show that the multi-task formulation presented in this paper is indeed effective in capturing non-linear task relationships and improves the learning performance on multi-task problems.

We have implemented this model with GPflow \citep{matthews_gpflow:_2017} and made our implementation publicly available on GitHub
\footnote{\url{https://github.com/aboustati/dgplib}}.

\section*{Acknowledgments}
This work was supported by The Alan Turing Institute under the EPSRC grant EP/N510129/1. AB gratefully acknowledges his scholarship award from the University of Warwick.

\bibliographystyle{abbrvnat}  
\bibliography{references}

\clearpage

\appendix
\onecolumn

\begin{center}
    \begin{huge}
    Supplementary Material
    \end{huge}
\end{center}

\input{sections/appendix_model_and_inference.tex}
\input{sections/appendix_experiment_details.tex}
\input{sections/appendix_mnist.tex}
\input{sections/appendix_sarcos.tex}
\end{document}

%% file: abbrev.tex
\newcommand*{\mtldgp}{mMDGP}
\newcommand*{\sharedmtldgp}{sMDGP}
\newcommand*{\icmdgp}{cMDGP}
\newcommand*{\stldgp}{iDGP}
\newcommand*{\stlgp}{iGP}
\newcommand*{\icmgp}{cGP}
\newcommand*{\stlrf}{iRF}
\newcommand{\mtlannTwo}{mANN2}
\newcommand{\mtlannThree}{mANN3}
\newcommand{\mtlbnnTwo}{mBNN2}
\newcommand{\mtlbnnThree}{mBNN3}

\newsavebox\MBox
\newcommand\Cline[2][red]{{\sbox\MBox{$#2$}%
  \rlap{\usebox\MBox}\color{#1}\rule[-2\dp\MBox]{\wd\MBox}{1.5pt}}}
  
\definecolor{green}{RGB}{0,128,0}

%% file: sections/introduction.tex
\section{INTRODUCTION}
Multi-task learning is a broad framework that aims to leverage the shared information between the training signals of related tasks in order to improve generalisation across them. This framework has been applied to various models such as neural networks \citep{caruana_multitask_1997}, support vector machines \citep{evgeniou2004regularized} and probabilitic models \citep{bonilla_multi-task_2007}. 

In probabilistic machine learning, a widely successful formulation of multi-task learning builds on the Gaussian process (GP) literature. In general, GPs \citep{rasmussen_gaussian_2006} provide a powerful and flexible non-parametric family of machine learning models, suitable for many nonlinear tasks. Their multi-task counterparts have also been widely adopted as a viable probabilistic alternative to classical multi-task learning models. However, the vast majority of GP multi-task models make a strong linearity assumption on the task dependencies, i.e. the tasks constitute a linear mixture of latent processes \citep{goovaerts_geostatistics_1997, whye_teh_semiparametric_2005, bonilla_multi-task_2007, alvarez_kernels_2011, wilson_gaussian_2012, nguyen_collaborative_2014}. Indeed, some attempts have been made to introduce non-linear task relationships to GP models \citep{boyle_dependent_2004, alvarez_sparse_2009, alaa_deep_2017, requeima2019gaussian}; however, at the expense of additional limiting assumptions or deviation from the desirable plug-and-play nature of GP models, e.g. \citet{alaa_deep_2017} and \citet{requeima2019gaussian} are restricted to dataset with fully observed outputs for all input locations. 

\begin{figure}[h]
\centering
\includegraphics[width=0.9\linewidth]{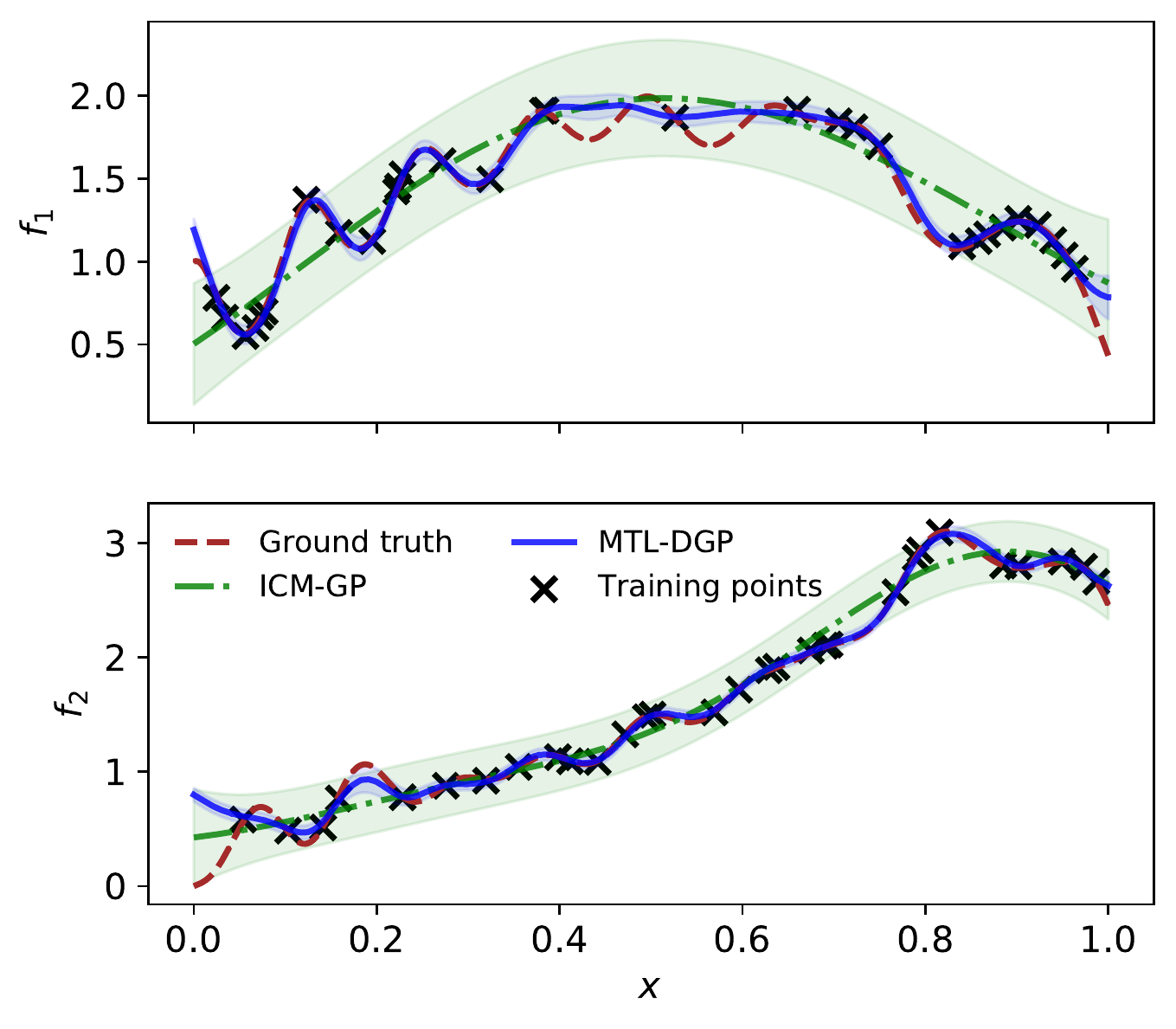}
\caption{Illustration of the fit of linear (ICM-GP) vs non-linear (MTL-DGP) multi-task GP models on a toy dataset with non-linear dependencies between the tasks. The ICM-GP model suffers from negative transfer due to its linear coregionalization assumption. The non-linear multi-task DGP is robust against it.}
\label{fig:example}
\end{figure}

To further highlight the limitation of linear mixing models, consider the following toy example with two tasks composed by non-linearly combining two private processes and a shared process:
\begin{align*}
    g(x) &= -\sin{(8\pi (x+1))}/(2x+1) - x^4, \nonumber\\
    h_1(x) &= \sin(3x), \qquad h_2(x) = 3x, \nonumber\\
    f_1(x) &= \cos^2{(g(x))} + h_1(x), \nonumber\\
    f_2(x) &= \sin{(10x)}g^2(x) + h_2(x),
\end{align*}
with $x \in [0, 1]$. Here, $f_1$ and $f_2$ are the two tasks, generated by $g$, $h_1$, $h_2$. Clearly, $f_1$ and $f_2$ have a complex non-linear relationship that linear mixing models struggle to capture as illustrated in \cref{fig:example}. 

To address this issue, propose a Deep Gaussian Process (DGP) approach to non-linear multi-task learning (illustrated in blue in \cref{fig:example}). DGPs a hierarchical composition of GPs that retain the advantages of GP-based models such as their non-parametric formulation, quantification of uncertainty and robustness to overfitting, while at the same time offering a richer class of models with the ability to learn complex representations from data. Our proposed framework can capture highly non-linear relationships between tasks, extending the popular linear construction, while retaining most of its flexibility.

Our modelling strategy assumes that similar tasks arise from a collection of underlying latent processes, some shared between the tasks and some that are task specific. These are combined \textit{non-linearly} using a further GP layer, enabling modelling of richer task relationships. Closed-form inference in the resulting model is intractable, thus we extend the doubly stochastic variational approximation in \citet{salimbeni_doubly_2017} to handle the multi-task DGP case. We demonstrate the capabilities of our model through experimentation on three datasets showing that the multi-task learning formulation for DGPs outperforms other single-task and multi-task GP-based models and neural networks on multiple benchmarks.

%% file: sections/model_specification.tex
\section{MODELLING APPROACH}
\subsection{MODELLING WITH GAUSSIAN PROCESSES}
GPs are used in Bayesian predictive modelling where the outputs are modelled as a transformation of an unknown function of the inputs $\{\mathbf{x}_n\}_{n=1}^N$, such that
\begin{equation*}
\mathbf{y}_n | f; \mathbf{x}_n \sim p(\mathbf{y}_n | f(\mathbf{x}_n)),
\end{equation*}
where $f$ is the latent function drawn from a GP prior $f(\cdot) \sim \mathcal{GP}(\mathbf{0}, k(\cdot, \cdot))$, and $p(\cdot)$ is an appropriate likelihood capturing the assumptions on the observations $\{\mathbf{y}_n\}_{n=1}^N$, e.g. Gaussian noise. 

The learning capacity for a GP model is primarily determined by the covariance function $k(\cdot, \cdot)$, which encodes the prior beliefs on the latent function $f$, as well as the mean function $m(\cdot)$ which encodes trends. For complex input-to-output mappings, one needs to select appropriate covariance functions that can capture the desired properties of such mappings, a process that requires extensive domain knowledge of the underlying problem. To introduce further flexibility into the modelling process, an alternative route is to compose multiple simple latent functions into an overall deep architecture
\begin{equation*}
\mathbf{y}_n | f^L; \mathbf{x}_n \sim p(\mathbf{y}_n | f^L(f^{L-1}(\dots f^1(\mathbf{x}_n)))),
\end{equation*}
with $f^l(\cdot) \sim \mathcal{GP}(m^l(\cdot), k^l(\cdot, \cdot))$. This model is known as a Deep Gaussian Process (DGP) \citep{damianou_deep_2013}. For all but the simplest cases, exact inference in DGPs is intractable, requiring approximate inference, e.g. by using the variational sparse approximation framework \citep{titsias_variational_2009, hensman_gaussian_2013}, extended to DGPs in \citet{salimbeni_doubly_2017}.

\subsection{DGP MULTI-TASK FORMULATION}
\label{model}

\begin{figure}[t]
\centering
\includegraphics[width=0.9\linewidth]{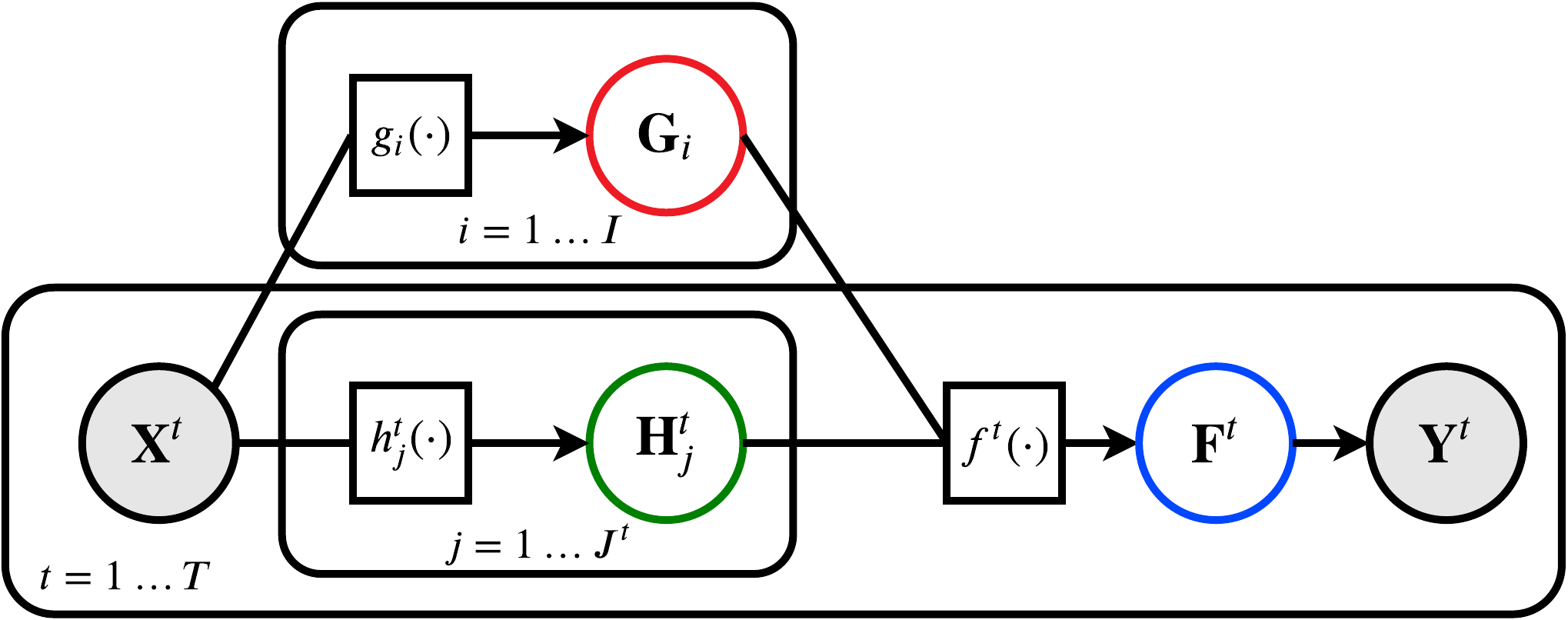}
\caption{A graphical representation of the proposed non-linear multi-task DGP model. The unshaded circles represent latent variables, the shaded circles represent observed variables and the squares are computational graph nodes. Some circles are colour coded to match their corresponding terms in \eqref{eqn:expanded_joint}, \eqref{eqn:approx_posterior} and \eqref{eqn:mtl-elbo}.}
\label{fig:model}
\end{figure}

Consider the case with $T$ tasks. Let $\mathbf{X}^t$ be the $N^t \times D^{\text{in}}$ data matrix for task $t \in \{1, \dots, T\}$ and $\mathbf{Y}^t$ be an $N^t \times D^{\text{out}}$ matrix corresponding to the outputs for task $t$. We propose a model where all $T$ tasks share a set of $I$ latent representations $\{\mathbf{G}_i\}_{i=1}^I$ generated by a set of $I$ functions $\{g_i(\cdot)\}_{i=1}^I$, where $g_i:\mathbb{R}^{D^\text{in}}\rightarrow\mathbb{R}^{D^{G_i}}$. In addition, they possess a set of $J_t$ task specific representations $\{\mathbf{H}_j^t\}_{j=1}^{J^t}$ generated by a set of $J^t$ task specific functions $\{h_j^t(\cdot)\}_{j=1}^{J^t}$ for each task $t$, where $h_j^t:\mathbb{R}^{D^\text{in}}\rightarrow\mathbb{R}^{D^{H_j^t}}$. The features in the combined latent space $\{\Lambda^t\}_{t=1}^T$ (with $\Lambda^t = \{\{\mathbf{G}_i\}_{i=1}^I, \{\mathbf{H}_j^t\}_{j=1}^{J^t}\}$) are warped with a task specific random function $f^t(\cdot)$ with $f^t:\mathbb{R}^{D^t}\rightarrow\mathbb{R}^{D^{\text{out}}}$ to generate noiseless outputs $\mathbf{F}^t$ for task $t$ (\cref{fig:model}).\footnote{$D^t = \sum_{i=1}^I D^{G_i}+\sum_{j=1}^J D^{H_j^t}$}

We place independent GP priors on both the shared and task specific latent functions:
\begin{align*}
g_i &\sim \mathcal{GP}(m_i(\mathbf{X}), k_i(\mathbf{X}, \mathbf{X}^{\prime})), \\
h_j^t &\sim \mathcal{GP}(m_j^t(\mathbf{X}^{t}), k_j^t(\mathbf{X}^{t}, \mathbf{X}^{t \prime})), \\
f^t | g, h^t &\sim \mathcal{GP}(m^t(\Lambda^t), k^t(\Lambda^t, \Lambda^t)).
\end{align*}
This generative process describes a 2-level hierarchical model with GPs that can be formulated as a 2-layer DGP, where the first layer is responsible for extracting shared and task specific features, while the second layer transforms the features into task specific outputs.

\paragraph{Sparse Variational Approximation} Exact inference in this model is intractable due to the non-linear dependencies introduced between the inner layers \citep{damianou_deep_2013} and between the tasks in the output layers. Hence, we resort to approximate inference where we extend the doubly stochastic variational sparse approximation framework in \citet{salimbeni_doubly_2017} to handle the multitask case.

Define the joint distribution of an expanded model as
\begin{align}
p(\{\mathbf{Y}^t, \mathbf{F}^t, \mathbf{\tilde{F}}^t, \{\mathbf{H}^t_j, \mathbf{\tilde{H}}^t_j\}_{j=1}^{J^t}\}_{t=1}^T, \{\mathbf{G}_i, \mathbf{\tilde{G}}_i\}_{i=1}^I) = &
[\prod_{t=1}^T \prod_{n=1}^{N^t}p(\mathbf{y}_n^t|\mathbf{F}^t)] [\prod_{t=1}^T \Cline[blue]{p(\mathbf{F}^t|\mathbf{\tilde{F}}^t, \Lambda^t) p(\mathbf{\tilde{F}}^t)}] \nonumber \\
                                         &[\prod_{t=1}^T\prod_{j=1}^{J^t} \Cline[green]{p(\mathbf{H}_j^t|\mathbf{\tilde{H}}_j^t) p(\mathbf{\tilde{H}}_j^t)}]
                                 [\prod_{i=1}^I \Cline[red]{p(\mathbf{G}_i|\mathbf{\tilde{G}}_i) p(\mathbf{\tilde{G}}_i)}],
\label{eqn:expanded_joint}
\end{align}
where we introduce the inducing variables $\{\mathbf{\tilde{G}}_i\}_{i=1}^I, \{\{\mathbf{\tilde{H}}^t_j\}_{j=1}^{J^t}\}_{t=1}^T, \{\mathbf{\tilde{F}}^t\}_{t=1}^{T}$ corresponding to the values of the latent functions evaluated at a set of $M$ inducing locations, i.e. $\mathbf{\tilde{G}}_i = g_i(\mathbf{Z}_{G_i})$, $\mathbf{\tilde{H}}^t_j = h^t_j(\mathbf{Z}_{\mathbf{H}_j^t})$, $\mathbf{\tilde{F}}^t = f^t(\mathbf{Z}_{\mathbf{F}^t})$. Augmenting the model this way allows factorising the joint GP priors into the prior on the inducing variables and the GP conditional given the inducing variables \citep{titsias_variational_2009}. For instance, in the case of the tuple $(\mathbf{G}_i, \mathbf{\tilde{G}}_i)$, we have
\begin{equation}
	p(\mathbf{G}_i, \mathbf{\tilde{G}}_i; \mathbf{X}, \mathbf{Z}_{\mathbf{G}_i}) = p(\mathbf{G}_i | \mathbf{\tilde{G}}_i ; \mathbf{X}, \mathbf{Z}_{\mathbf{G}_i}) p(\mathbf{\tilde{G}}_i; \mathbf{Z}_{\mathbf{G}_i}),
\end{equation}
where
\begin{align}
&p(\mathbf{\tilde{G}}_i; \mathbf{Z}_{\mathbf{G}_i}) = \mathcal{N}(\mathbf{\tilde{G}}_i | m_i(\mathbf{Z}_{\mathbf{G}_i}), k_i(\mathbf{Z}_{\mathbf{G}_i}, \mathbf{Z}_{\mathbf{G}_i})), \\
&p(\mathbf{G}_i | \mathbf{\tilde{G}}_i ; \mathbf{X}, \mathbf{Z}_{\mathbf{G}_i}) = \mathcal{N}(\mathbf{G}_i |  \boldsymbol{\tilde{\mu}_i}, \boldsymbol{\tilde{\Sigma}}_i), 
\end{align}
with
\begin{align}
&\boldsymbol{\tilde{\mu}}_i = m_i(\mathbf{X}) + \boldsymbol{\alpha}_i(\mathbf{X})^T(\mathbf{\tilde{G}}_i-m_i(\mathbf{Z}_{\mathbf{G}_i})), \\
&\boldsymbol{\tilde{\Sigma}}_i = k_i(\mathbf{X}, \mathbf{X}) - \boldsymbol{\alpha}_i(\mathbf{X})^T k_i(\mathbf{Z}_{\mathbf{G}_i}, \mathbf{Z}_{\mathbf{G}_i}) \boldsymbol{\alpha}_i(\mathbf{X}), \\
&\boldsymbol{\alpha}_i(\mathbf{X}) = k_i(\mathbf{Z}_{\mathbf{G}_i}, \mathbf{Z}_{\mathbf{G}_i}) ^{-1} k_i(\mathbf{Z}_{\mathbf{G}_i}, \mathbf{X}).
\end{align}
Similarly for $(\mathbf{H}_j^t, \mathbf{\tilde{H}}_j^t)$ and $(\mathbf{F}^t, \mathbf{\tilde{F}}^t)$ taking the input pairs $(\mathbf{X}^t, \mathbf{Z}_{\mathbf{H}_j^t})$ and $(\Lambda^t, \mathbf{Z}_{\mathbf{F}^t})$ respectively.

Expanding the probability space this way allows us to define an approximate variational posterior
\begin{align}
q(&\{\mathbf{F}^t, \mathbf{\tilde{F}}^t, \{\mathbf{H}^t_j, \mathbf{\tilde{H}}^t_j\}_{j=1}^{J^t}\}_{t=1}^T, \{\mathbf{G}_i, \mathbf{\tilde{G}}_i\}_{i=1}^I) = \nonumber \\
& [\prod_{t=1}^T \Cline[blue]{p(\mathbf{F}^t|\mathbf{\tilde{F}}^t, \Lambda^t) q(\mathbf{\tilde{F}}^t)}]
                                         [\prod_{t=1}^T\prod_{j=1}^{J^t} p\Cline[green]{(\mathbf{H}_j^t|\mathbf{\tilde{H}}_j^t) q(\mathbf{\tilde{H}}_j^t)}] 
                                [\prod_{i=1}^I \Cline[red]{p(\mathbf{G}_i|\mathbf{\tilde{G}}_i) q(\mathbf{\tilde{G}}_i)}],
\label{eqn:approx_posterior}
\end{align}
where the approximate posteriors $q$'s on the right-hand side are free-form Gaussians, e.g. $q(\mathbf{\tilde{G}}_i) = \mathcal{N}(\mathbf{\tilde{G}}_i| \mathbf{m}_i, \mathbf{G}_i)$ and similarly for $\mathbf{\tilde{H}}_j^t$ and $\mathbf{\tilde{F}}^t$. 

\paragraph{Evidence Lower Bound (ELBO)} We can derive an ELBO
\begin{align}
\displaystyle
\mathcal{L} =& \sum_{t=1}^T \sum_{n=1}^{N^t} \mathbb{E}_{q(\mathbf{f}_n^t)}[\log p(\mathbf{y}_n^t | \mathbf{f}_n^t)] 
-\sum_{t=1}^T \Cline[blue]{\text{KL}\left[q(\mathbf{\tilde{F}}^t) \middle\| p(\mathbf{\tilde{F}}^t)\right]} \nonumber \\
&-\sum_{t=1}^T \sum_{j=1}^{J^t} \Cline[green]{\text{KL}\left[q(\mathbf{\tilde{H}}^t_j) \middle\| p(\mathbf{\tilde{H}}_j^t)\right]} -\sum_{i=1}^I \Cline[red]{\text{KL}\left[q(\mathbf{\tilde{G}}_i) \middle\| p(\mathbf{\tilde{G}}_i)\right]},
\label{eqn:mtl-elbo}
\end{align}
by taking the expected log-ratio of the \eqref{eqn:expanded_joint} and \eqref{eqn:approx_posterior} (detailed derivation in \cref{sec:detailed-elbo}). This bound has complexity $\mathcal{O}(NM^2(\sum_{i=1}^I D^{G_i} + \sum_{t=1}^T \sum_{j=1}^{J^t}D^{H_j^t} + D^\text{out}))$ and can be evaluated approximately using Monte Carlo samples from the variational posterior.

\paragraph{Do we need deeper architectures?} The description of the model is presented for a 2-layer DGP; however, this framework is general and can be applied to a DGP with arbitrary depth. Looking at the ELBO in \eqref{eqn:mtl-elbo}, the contribution of the first layer appears in the KL terms involving the $\mathbf{\tilde{G}}_i$'s and the $\mathbf{\tilde{H}}_j^t$'s. Therefore, a layer of this structure can, in principle, be added at any level of a DGP cascade by simply adapting the ELBO to include these terms. Since the sparse approximation and the inference framework in \citet{salimbeni_doubly_2017} do not require the computation of the full covariance within the layers, it is possible to propagate a Monte Carlo sample through this layer structure at any point in the cascade without the need to compute inter-task covariances. This enables approximate evaluation of the expected log-likelihood term in \eqref{eqn:mtl-elbo}. 

It is worth noting, however, that increasing the depth this way adds extra complexity to the model and potentially violates the generative assumption of the non-linear combination of shared and task specific processes. Hence, we recommend that such extension should only be considered when the structure of the problem requires it.

\subsection{MODEL SPECIFICATION}
\label{sec:model_instantiation}
So far, we have described the modelling approach in an unspecified setting. One can use to the proposed framework as a base to construct a plethora of multi-task DGP models. In this section, we will discuss three straightforward instantiations of this framework.

\paragraph{The multiprocess Multi-task DGP (\mtldgp{})} follows the formulation in \cref{model} closely. The model consists of a two layer GP with the inner layer (latent space) split into a set of independent shared processes and independent task specific processes. The outputs of those processes are concatenated and fed into the top layer which constitute independent task specific GPs. The GPs in the top layer admit Automatic Relevance Determination (ARD) kernels \citep{rasmussen_gaussian_2006}
\begin{equation}
    k(\lambda, \lambda^\prime) = \sigma^2 \kappa\big(\sum_{i=1}^{D^\lambda}\omega_k s(\lambda_i - \lambda_i^\prime)\big),
\end{equation}
where $\kappa(\cdot)$ is a positive definite function, $s(\cdot)$ is a distance function and $\sigma$ and $\omega_i$ are the kernel parameters, which are learned by maximising the ELBO as a surrogate to the model's marginal likelihood. 

The use of the ARD kernel in the top layer enables each output process to weigh the information from the latent processes differently, thus balancing the use of shared information and task specific information (see \cref{fig:hinton} in \cref{sec:sarcos} for illustration). This can guard against some types of negative transfer. For instance, if a contaminating task is present in the dataset, the model can, in principle, learn ARD weights such that all its information are absorbed in the task specific components. 

\paragraph{The shared Multi-task DGP (\sharedmtldgp{})} consists of a 2-layer DGP where all the latent processes are shared between the tasks with no task specific processes. It also employs an ARD kernel in the top layer. The role of ARD in this version is akin to Manifold Relevance Determination \citep{damianou_manifold_2012}, where the ARD weights determine a soft segmentation of the latent space.

\paragraph{The Coregionalised Multi-task DGP (\icmdgp{})} is similar to \sharedmtldgp{}, with a single shared latent space. The middle layer forms a collection of coregionalised GPs \citet{goovaerts_geostatistics_1997, alvarez_kernels_2011}, i.e. unlike \sharedmtldgp{} the outputs of the processes in the latent space are not concatenated but linearly mixed with different mixing coefficients for different tasks. In addition, the top layer is set to a single coregionalised GP that outputs all the tasks. This formulation is similar to \citet{alaa_deep_2017}, used for survival analysis with competing risks. However, the inference scheme presented in \citet{alaa_deep_2017} is only applicable with certain covariance functions and is not suitable for the \emph{asymmetric} multi-task learning case (where only a subset of the outputs is observed for a single input location). Conversely, using the doubly stochastic formulation for the ELBO allows using any valid covariance function and enables the inference on partially observed outputs.

%% file: sections/related_work.tex
\section{RELATED WORK}
There is a rich literature on multi-task learning in GP models. Most GP multi-task models correlate the outputs by mixing a set of independent processes with a different set of coefficients for each output. Examples of such models include: the semi-parametric latent factor model (SLFM) \citep{whye_teh_semiparametric_2005}, intrinsic coregionalisation model (ICM) \citep{bonilla_multi-task_2007,skolidis_bayesian_2011} and the linear model for coregionalisation (LMC) \citep{goovaerts_geostatistics_1997,alvarez_kernels_2011}. \citet{nguyen_collaborative_2014} extend SLFM to handle dependent tasks in very large datasets using the framework in \citet{hensman_gaussian_2013}. In \citet{titsias_spike_2011}, the authors give this type of models a Bayesian treatment by placing a spike-and-slab prior on the mixing coefficients, and in \citet{aglietti19a}, the authors treat the coefficients as stochastic processes indexed by task discriptors in a spatial Cox process setting. 

More complex models that can handle complex relationships between the outputs are formulated in \citet{wilson_gaussian_2012} and \citet{nguyen_efficient_2013} introducing mixing weights with input dependencies, and in \citet{boyle_dependent_2004} and \citet{alvarez_sparse_2009} by convolving processes. 

In the DGP literature, there are two notable formulations for the multi-task setting. \citet{kandemir_asymmetric_2015} proposes to linearly combine the hidden layers of different DGP models trained on different tasks, inducing information transfer between the models. \citet{alaa_deep_2017} uses the ICM kernel in the hidden and output layers of the DGPs for multi-task learning for survival analysis with competing risks. Our framework is a general formulation of the models above, replacing the linear mixing of latent processes with another GP, thus allowing to model more complex relationships between the tasks. The work of \cite{requeima2019gaussian} on the Gaussian Process Autoregressive Regression (GPAR) model offers an alternative formulation to non-linear multi-output learning in GPs, This model assumes an inherit ordering of the outputs and places independent GP priors on them, where the current GP output is concatenated with the inputs of the proceeding GP at the observed location. This model can be interpreted as a particular DGP structure with skip connections. However, in the multi-task setting this model assumes an inherit ordering of the tasks and that the dataset is closed downwards, i.e. for task $t$, all the previous $t-1$ outputs are observed for each input location in the dataset; otherwise this can be remedied with ad-hoc imputation of the missing data.

A similar idea to multi-task learning is multi-view learning, where multiple views of the data are combined into a single latent representation. In the Gaussian process literature, Manifold Relevance Determination (MRD) \citep{damianou_manifold_2012} is a latent variable model for multi-view learning, where the latent space is softly separated into a component that is shared between all of the views and private components that are view specific. In this work, we attempt to do the inverse mapping, i.e. disentangling multiple representations from a common feature space. In \sharedmtldgp{}, we use the idea of the soft separation in the latent space, and in \mtldgp{} we encode a hard separation between private components and task-specific components of the latent space.

We briefly explore the connection of the proposed model to some of the above models and other models in the wider multi-task learning literature.

\subsection{LINEAR PROCESS MIXING}
To make the relationship of the proposed model to other GP-based models more explicit, we consider the following construction of a multi-output function $f(x)$ of $T$ outputs, where we refer to the $t$th output as
\begin{equation}
    f^t(\mathbf{x}) = \sum_{i=1}^I w^t_i g_j(\mathbf{x}) + \sum_{j=i}^J h^t_j(\mathbf{x}),
    \label{eqn:linear-basis}
\end{equation}
where $\{w_i^t\}_{i=1}^I$ is a set of weights associated with output $t$, and the $g_j(\cdot)$'s and $h^t_j(\cdot)$'s are basis functions that are shared and output specific, respectively. In the probabilistic literature, the set of basis functions are modelled as independent GPs with their choice of covariance function determining the particular model, e.g. setting independent GP priors on the basis functions with different covariance functions recovers the Semi-parametric Latent Factor Model of \citet{whye_teh_semiparametric_2005}. Since the output processes are a sum of GPs, it follows that they themselves are also GPs with a tractable cross-covariance that depends on the weights on the shared basis functions.

We can generalise the construction in \eqref{eqn:linear-basis} to
\begin{equation}
    f^t(\mathbf{x}) = \varphi^t(g_1(\mathbf{x}), \dots, g_I(\mathbf{x}), h_1^t(\mathbf{x}), \dots, h_J^t(\mathbf{x})),
    \label{eqn:composition}
\end{equation}
where $\varphi^t(\cdot)$ is an arbitrary function. Depending on the choice of $\varphi^t(\cdot)$, this construction combines the basis functions non-linearly; for instance, allowing interactions between the shared and output specific processes. Treating this construction probabilistically, we can place GP priors on the basis function as well as the warping function $\varphi^t(\cdot)$, which recovers the proposed model. It is important to note that due to the non-linearity of $\varphi^t(\cdot)$, the resulting output processes are no longer Gaussian and their cross-covariance cannot be characterised in terms of the shared basis functions only. As a side note, choosing a GP with linear covariance function as a prior on $\varphi^t(\cdot)$ recovers the linear model described in \eqref{eqn:linear-basis}.

\subsection{PROCESS CONVOLUTION}
One can formulate a GP as the convolution of a base process with a smoothing kernel, i.e.
\begin{equation}
    f(\mathbf{z}) = \int_\chi G(\mathbf{x} - \mathbf{z})u(\mathbf{z})d\mathbf{z},
    \label{eqn:convolution}
\end{equation}
where the smoothing kernel $G(\cdot)$ is square integrable (i.e. $\int G^2(x-u)du < \infty$) and the base process $u(\cdot)$ is a white noise process, or more generally any random process (although if non-Gaussian then the resulting convolution process is not a GP) \citep{calder2007some}. In the multi-output process convolution literature, different outputs are assumed to share the same base process $u$, with different smoothing kernels for each output. Hence, the cross-covariance between the output processes can be derived tractably as convolution is a linear operation.

Using \eqref{eqn:convolution} and conditioned on the latent processes, we can write the top layer of the presented model as
\begin{equation}
    f^t(\mathbf{x}) = \int_\chi G^t(\mathbf{g} - \mathbf{z}_g, \mathbf{h}^t - \mathbf{z}_h)u^t(\mathbf{z})d\mathbf{z},
    \label{eqn:mtl-convolution1}
\end{equation}
assuming a separable smoothing kernel $G^t(\mathbf{g} - \mathbf{z}_g, \mathbf{h}^t - \mathbf{z}_h) = G^t(\mathbf{g} - \mathbf{z}_g) G^t( \mathbf{h}^t - \mathbf{z}_h)$ (which is the case for processes with ARD kernels), we can rewrite \eqref{eqn:mtl-convolution1} as
\begin{align}
    f^t(\mathbf{x}) =& \int_\chi G^t(\mathbf{g} - \mathbf{z}_g)G^t( \mathbf{h}^t - \mathbf{z}_h)u^t(\mathbf{z})d\mathbf{z} \nonumber \\
    =& \int_\chi G^t(\mathbf{g} - \mathbf{z}_g)\upsilon^t(\mathbf{z}_g)d\mathbf{z}_g,
    \label{eqn:mtl-convolution2}
\end{align}
where $\upsilon^t$ is also random process. Since by construction, $g$ itself is a random (Gaussian) process, we can see that the output correlation is induced in the smoothing kernel rather than the base process. Conditioning the smoothing kernel on a random process is explored in \citet{higdon1998process} and \citet{higdon1999non} to induce non-stationary behaviour in the output process. To the best of our knowledge this work is the first instance to consider this conditioning for information sharing behaviour.

\subsection{REGULARISATION METHODS}
The ELBO defined in \eqref{eqn:mtl-elbo} has a similar formulation to the general multi-task objective function in regularisation-based multi-task learning literature \citep{evgeniou2004regularized}. A general formulation of which is given by
\begin{equation}
    \label{eqn:mtl-objective}
    L = \frac{1}{T} \sum_{t=1}^T L^t(\theta, \phi; \mathbf{Y}^t) + \Omega(\theta) + \sum_{t=1}^T\Psi(\phi^t),
\end{equation}
where $\theta$ are shared parameters, $\phi = \{\phi^t\}_{t=1}^T$ are task specific parameters and $\{L\}_{t=1}^T$, $\Omega$ and $\{\Psi\}_{t=1}^T$ are the task specific losses, shared regulariser and task specific regularisers respectively. The choice and strength of regularisation controls the information transfer between tasks.

A direct analogy can be drawn from \eqref{eqn:mtl-elbo} to \eqref{eqn:mtl-objective}, where the expected log-likelihood terms correspond to the loss functions and the KL terms correspond to the regularisers, respectively. This forms a new link between the probabilistic and the regularisation points-of-view of multi-task learning and motivates the prospect of choosing alternative prior processes to induce desirable behaviour in the model, e.g. sparsity in the latent space \citep{argyriou2007multi}. Furthermore, one can also frame \eqref{eqn:mtl-elbo} as a Generalised Variational Inference objective  \citep{knoblauch2019generalized}, where the KL terms are viewed as prior regularisers. Changing the form of these regularisers can induce different information sharing behaviour in the approximate predictive posteriors of the tasks. Unfortunately, exploring this point is beyond the scope of this work.

%% file: sections/mnist_variations.tex
\label{sec:mnist}
\begin{figure}[!ht]
\centering
\includegraphics[width=0.6\linewidth]{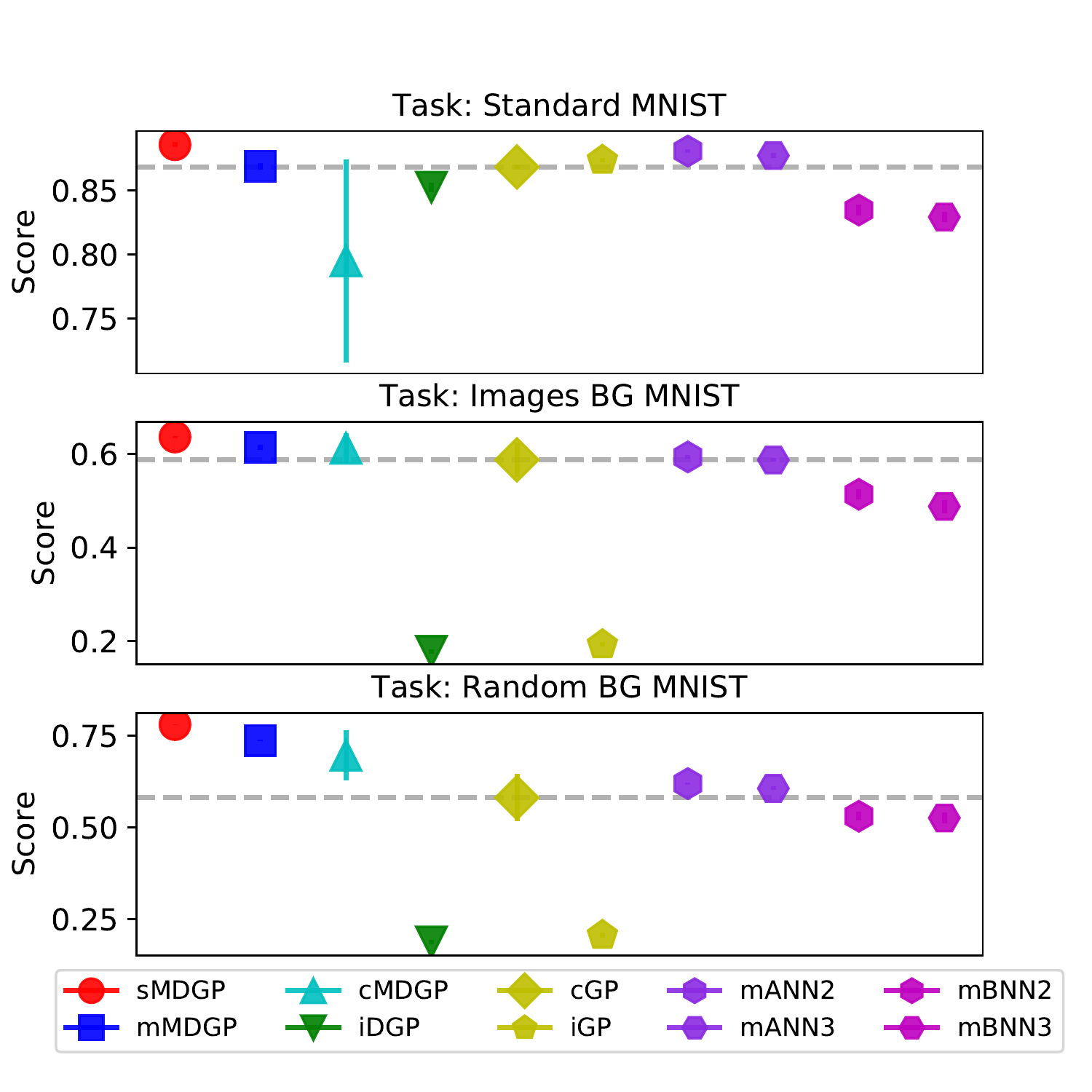}
\caption{Average classification accuracy and its standard error on the MNIST variations experiments. The dashed line highlights the multi-task GP baseline. Higher is better.}
\label{fig:mnist}
\end{figure}

\subsection{MNIST VARIATIONS}
We use the MNIST digits dataset \citep{lecun_gradient-based_1998}, as well as two other variations. The first variation,  \textbf{Images Background (BG) MNIST}, replaces the background with randomly extracted patches from 20 black and white images. The second variation, termed termed \textbf{Random Background (BG) MNIST}, replaces the black background for the digits with random noise \footnote{\url{https://sites.google.com/a/lisa.iro.umontreal.ca/public_static_twiki/variations-on-the-mnist-digits}}. These two variations are more challenging for classification due to the presence of non-monochromatic backgrounds.

We sample 1000 images from the training split each of the three datasets. Hence, we have 3000 training data points in total for 3 multi-class classification tasks. We report the average accuracy and its standard error on the standard test split for the datasets over 10 runs. The purpose of this experiment is to test the ability of the model to transfer information from a relatively easy task, standard MNIST classification, to more challenging tasks, i.e. images BG MNIST and random BG MNIST.

Figure \ref{fig:mnist} suggests that MDGP models are successfully able to use common information shared between the three tasks to improve the learning performance on the difficult tasks (image BG MNIST and random BG MNIST). However, we see that \icmdgp{} suffers from negative transfer on the the first task, due to its inability to weigh the contributions of the latent processes differently for each task. In contrast, \mtldgp{} and \sharedmtldgp{} do not suffer from this problem, highlighting their robustness to this type of negative transfer from the use of ARD weights on the warping layer. 

MDGP models achieve a considerable performance gain on the third task (random BG MNIST) compared to all the other models, by a factor of almost 50\%. This task is the most challenging out of the three as the background here is not structured. This improvement highlights the importance of the non-linear mixing of the latent processes to deal with complex task structures.

\begin{figure}[!h]
\centering
\includegraphics[width=0.7\linewidth]{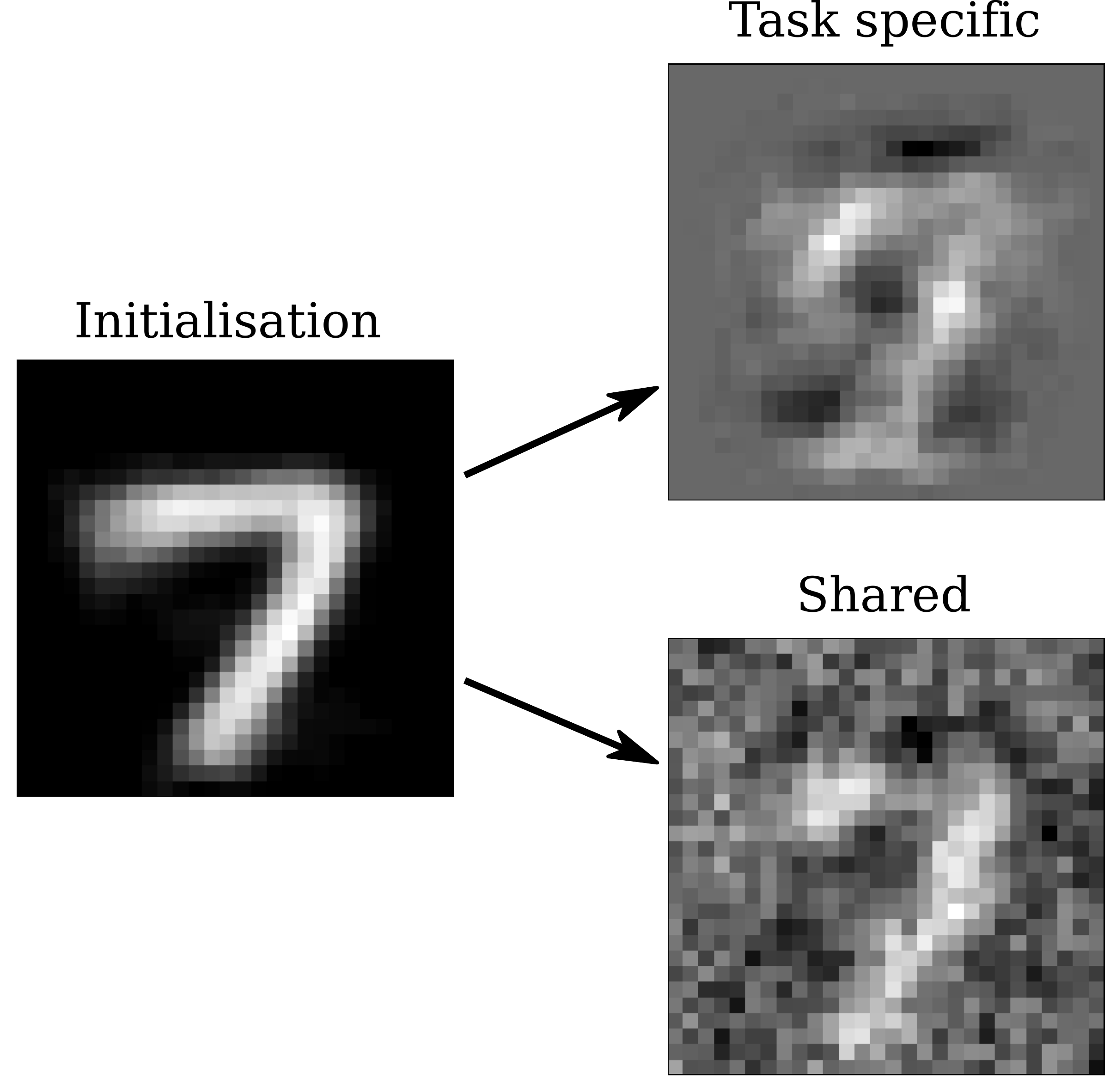}
\caption{Example of the learning behaviour of \mtldgp{}. The image on the left is the initial value for one of the inducing locations. The top right image is the learned inducing location for the task-specific layer for the Standard MNIST task. The bottom right image is the learned inducing location for the shared layer. The background information is encoded in the task specific representation, where the digit is blurred The shared information contains the digit on a unstructured background.}
\label{fig:mnist-illustration}
\end{figure}

To illustrate the learning behaviour of \mtldgp{}, we examine the inducing inputs from the inner layer. As specified in \cref{sec:model_instantiation}, \mtldgp{} has an inner layer with hard separation between the task-specific and shared processes. We made this modelling choice expecting the task specific part to learn private task information, while the shared part would learn global information shared across the tasks. \cref{fig:mnist-illustration}, obtained from one of the experiment runs, confirms this assumption. On the left we can see the initial value of an inducing location before optimisation. The optimisation transforms the initialisation in the task-specific component and the shared component of the inner layer in different ways. The task-specific component encodes the background information specific to the task, blurring the digit. The shared component, on the other hand, retains the structure of the digit removing background information.

%% file: sections/sarcos.tex
\subsection{SARCOS ROBOT INVERSE DYNAMICS}
\label{sec:sarcos}
\begin{table*}[ht]
\caption{Average NLPP scores on the SARCOS dataset over 7 tasks. The figures presented are the mean score (and standard error) over 10 runs. The lowest statistically significant scores based on a Wilcoxon test are presented in boldface. Lower is better.}
\vspace{.1in}
\centering
\begin{tabular}{lccccccc}  
\toprule
& \multicolumn{6}{c}{Number of Training Inputs} \\
\cmidrule(r){2-7}
 & 100 & 200 & 500 & 1000 & 2000 & 5000\\
\midrule
\sharedmtldgp & $1.47(0.26)$ & $\mathbf{0.98(0.11)}$ & $0.74(0.05)$ & $\mathbf{0.15(0.1)}$ & $\mathbf{-0.01(0.1)}$ & $\mathbf{-0.16(0.1)}$  \\
\mtldgp & $\mathbf{1.34(0.21)}$ & $1.05(0.10)$ & $\mathbf{0.67(0.12)}$ & $\mathbf{0.16(0.10)}$ & $\mathbf{0.01(0.11)}$ & $-0.12(0.10)$  \\
\icmdgp & $\mathbf{1.32(0.21)}$ & $\mathbf{0.95(0.1)}$ & $0.72(0.06)$ & $0.24(0.11)$ & $0.05(0.10)$ & $-0.09(0.10)$  \\
\stldgp & $\mathbf{1.32(0.17)}$ & $\mathbf{0.99(0.11)}$ & $0.75(0.06)$ & $0.64(0.04)$ & $0.36(0.09)$ & $0.02(0.08)$  \\
\icmgp & $1.51(0.06)$ & $1.42(0.04)$ & $1.35(0.04)$ & $1.27(0.08)$ & $0.12(0.11)$ & $0.02(0.11)$  \\
\stlgp & $1.43(0.03)$ & $1.38(0.02)$ & $1.34(0.03)$ & $1.28(0.07)$ & $1.17(0.09)$ & $0.91(0.12)$  \\
\mtlannTwo & $28.22(5.39)$ & $10.56(1.69)$ & $3.20(0.53)$ & $1.66(0.33)$ & $1.26(0.30)$ & $1.16(0.31)$  \\
\mtlannThree & $21.72(4.20)$ & $9.53(1.70)$ & $3.14(0.51)$ & $1.92(0.33)$ & $1.48(0.34)$ & $1.42(0.33)$  \\
\mtlbnnTwo & $1.58(0.22)$ & $1.11(0.20)$ & $\mathbf{0.55(0.09)}$ & $0.31(0.06)$ & $0.16(0.04)$ & $0.05(0.03)$  \\
\mtlbnnThree & $1.71(0.28)$ & $1.18(0.18)$ & $\mathbf{0.63(0.09)}$ & $0.37(0.06)$ & $0.21(0.04)$ & $0.09(0.03)$  \\
\bottomrule
\end{tabular}
\label{table:nlpp}
\end{table*}
We consider the SARCOS regression dataset relating to the inverse dynamics problem for a seven degrees-of-freedom anthropomorphic robot arm \citep{vijayakumar_statistical_2002,rasmussen_gaussian_2006}. The dataset consists of 44,484 training observations with 21 input variables and 7 outputs (tasks). Additionally, there are 4,449 testing examples with all the 7 outputs available.

The experimental procedure is described as follows: For $N$ in $\{100, 200, 500, 1000, 2000, 5000\}$, sample $N$ training data points from the training set. This constitutes the training set for one experimental run. For each of the $N$ sampled points select 1 of the 7 outputs uniformly at random. Split the experimental training set into 7 according to which joint the label corresponds to. These 7 splits constitute 7 tasks. We train the models on the training set after standardising the features and the targets and test for all 7 labels on the test set. We report the negative log predictive probability (NLPP) and the root mean squared error (RMSE), aggregated across the 7 tasks over averaged 10 runs (with standard errors). We use the NLPP scores to assess the models' ability to quantify uncertainty, and the RMSE scores to measure the mean predictions. The results for this experiment are presented in \cref{table:nlpp} and \cref{table:rmse} (in the appendix). 

Looking at the NLPP scores in \cref{table:nlpp}, we observe that for all data regimes MDGP models perform well, outperforming the shallow models for all configurations. As the dataset size increases the MDGP models are able to outperform the rest by a wider margin. The gain in performance of \icmdgp{} is not as pronounced as for \mtldgp{} and \sharedmtldgp{}, highlighting the importance of task specific warping functions with ARD kernels for this problem.

\begin{figure}[!ht]
\centering
\includegraphics[width=0.6\linewidth]{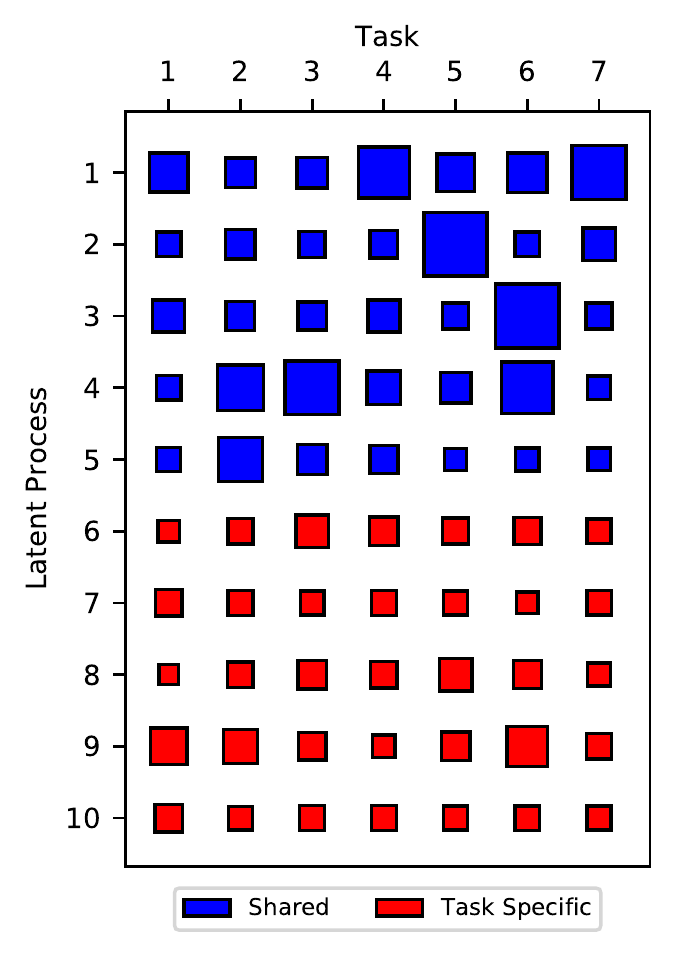}
\caption{Hinton diagram showing the ARD weights for on of the MDGP runs on Sarcos with 5000 training datapoints. Blocks in blue are shared latent processes, while blocks in red are task specific. This indicates that the model assigns different weights for the latent processes for each task.}
\label{fig:hinton}
\end{figure}

To take a closer look at the effect of the ARD kernel, \cref{fig:hinton} shows a Hinton diagram of the ARD weights for one run of \mtldgp{} on 5000 datapoints. The hard separation of latent processes is highlighted in the colour scheme. We can see that the model assigns different weights on the random processes for different tasks. It also on average assigns more weight to the shared processes than the task specific processes, indicating that the model utilises the shared representation. A similar diagram for \sharedmtldgp{} is shown \cref{fig:hinton2} in the appendix.

%% file: sections/faims.tex
\subsection{FAIMS DIABETES DIAGNOSIS}
\label{sec:faims}
Field Asymmetric Ion Mobility Spectrometry (FAIMS) is a method for detecting Volatile Organic Compounds (VOCs) which for example contribute to a given odour. VOCs are known to carry information on a range of disease states in humans, and technologies such as FAIMS can be used to cheaply and non-invasively diagnose a patient's disease from a simple biological sample such as urine \citep{covington_application_2015}.

We have access to data from a case-control study of 125 patients who have been tested for diabetes. 48 out of 125 have been found to have diabetes (the disease group), while the rest are disease-free (the control group). The data consists of three experimental runs per patient, corresponding to sequential FAIMS analyses on the same urine sample. These are expected to contain similar but not identical signals, as different VOCs evaporate at different rates, meaning the overall VOC signal varies over time. We treat each experimental run as a task, i.e. three binary classification tasks in total. By doing so, we aim to share statistical strength between runs to improve the accuracy of prediction and overcome the scarcity of training data.

Due to the large size of the feature space and its sparsity, we perform Sparse Principal Component Analysis decomposition \citep{hastie_elements_2009,mairal_online_2009} on the features selecting the first 20 principal components. This pipeline is standard for this type of datasets \citep{martinez-vernon_improved_2018}. We perform 10-fold cross-validation on 70:30 train-test splits. We report the value of the area under the Receiver Operating Characteristic Curve (ROC-AUC) for each task averaged across the folds, as well as the average ROC-AUC for all tasks and runs and their standard errors. We include a single task Random Forest classifier \citep{hastie_elements_2009} as a benchmark, which has been shown to perform well on this problem \citep{martinez-vernon_improved_2018}. The results of this experiment are shown in \cref{table:faims}.

We observe that MDGP models outperform the other GP-based models. They also significantly outperform the \stlrf{} baseline on the second task. Interestingly, both \stldgp{}  and \icmgp{} outperform \stlgp{} on average, indicating that representation learning as well as information sharing are important for this problem. MDGP models are able to leverage information transfer and nonlinear projection to perform well on this type of problems compared to the other tested models.
\begin{table}[bt]
\caption{ROC-AUC results on the FAIMS dataset averaged over 10 runs. The figures in parentheses are the standard errors. Higher is better.}
\vspace{.1in}
\label{table:faims}
\centering
\begin{tabular}{lrrrr}  
\toprule
& \multicolumn{4}{c}{ROC-AUC: mean(standard error)} \\
\cmidrule(r){2-5}
 Model & Task 1 & Task 2 & Task 3 & All \\
\midrule
\sharedmtldgp & $0.73(0.03)$& $0.80(0.02)$& $\mathbf{0.8(0.02)}$& $\mathbf{0.78(0.02)}$ \\
\mtldgp & $\mathbf{0.75(0.03)}$& $\mathbf{0.81(0.02)}$& $\mathbf{0.8(0.02)}$& $\mathbf{0.78(0.02)}$ \\
\icmdgp & $0.72(0.02)$& $\mathbf{0.81(0.02)}$& $0.79(0.02)$& $0.77(0.01)$ \\
\stldgp & $0.62(0.02)$& $0.75(0.02)$& $0.76(0.03)$& $0.71(0.02)$ \\
\icmgp & $0.58(0.03)$& $0.64(0.02)$& $0.66(0.02)$& $0.63(0.02)$ \\
\stlgp & $0.57(0.02)$& $0.56(0.03)$& $0.61(0.04)$& $0.58(0.02)$ \\
\stlrf & $0.71(0.03)$& $0.71(0.02)$& $0.72(0.02)$& $0.71(0.01)$ \\
\mtlannTwo & $0.54(0.03)$& $0.54(0.02)$& $0.50(0.03)$& $0.53(0.02)$ \\
\mtlannThree & $0.54(0.01)$& $0.52(0.03)$& $0.53(0.03)$& $0.53(0.02)$ \\
\mtlbnnTwo & $0.47(0.02)$& $0.48(0.02)$& $0.43(0.03)$& $0.46(0.01)$ \\
\mtlbnnThree & $0.44(0.04)$& $0.48(0.04)$& $0.45(0.02)$& $0.45(0.02)$ \\
\bottomrule
\end{tabular}
\end{table}

%% file: sections/appendix_model_and_inference.tex
\section{EXTRA TECHNICAL DETAILS}
\subsection{MODELLING WITH GAUSSSIAN PROCESSES}
GPs can be used in predictive modelling where the labels are modelled as a transformation of a non-parametric latent function of the inputs \citep{rasmussen_gaussian_2006}
\begin{equation*}
\mathbf{y}_n | f; \mathbf{x}_n \sim p(\mathbf{y}_n | f(\mathbf{x}_n)),
\end{equation*}
where $f$ is a latent function drawn from a GP which is usually set to be zero-mean, i.e. $f(\cdot) \sim \mathcal{GP}(\mathbf{0}, k(\cdot, \cdot))$, and $p(\cdot)$ is an appropriate likelihood, e.g. Gaussian or Bernoulli. Exact inference in this model is possible only when the likelihood is Gaussian and with computational complexity of $O(N^3)$. Therefore, practitioners often resort to a sparse variational approximation that allows the use of other likelihoods, as well as a reduction in computational complexity to $O(NM^2)$ \citep{titsias_variational_2009, hensman_gaussian_2013}, where $M$ is the number of inducing locations (pseudo inputs) which is typically much smaller than $N$.

The sparse variational approximation seeks to approximate the true GP posterior $p$, with an approximate posterior $q$, by minimising the Kullback-Leibler (KL) divergence between $q$ and $p$. This is equivalent to maximising a lower bound on the marginal likelihood of the model, known as the evidence lower bound (ELBO)
\begin{equation}
\mathcal{L} = \mathbb{E}_{q(\mathbf{F}, \mathbf{U})}\left[\log \frac{p(\mathbf{Y}, \mathbf{F}, \mathbf{U})}{q(\mathbf{F}, \mathbf{U})}\right],
\label{eqn:gp_elbo}
\end{equation}
where $\mathbf{Y} = [\mathbf{y}_1, \dots, \mathbf{y}_N]$ are the outputs,  $\mathbf{F} = [f(\mathbf{x}_1), \dots, f(\mathbf{x}_N)]$ are the latent function values and $\mathbf{U} = [f(\mathbf{z}_1), \dots, f(\mathbf{z}_M)]$ are the function values at the inducing locations $\mathbf{z}_m$.

In addition to allowing for more scalable inference in standard GP models, the sparse variational approximation enables tractable inference in models involving the composition of GPs; for instance DGPs. Recall, a DGP is a composition of functions with GP priors on each and i.i.d Gaussian noise between the layers \citep{damianou_deep_2013}
\begin{equation*}
\mathbf{y}_n | f^L; \mathbf{x}_n \sim p(\mathbf{y}_n | f^L(f^{L-1}(\dots f^1(\mathbf{x}_n)))),
\end{equation*}
where $f^l(\cdot) \sim \mathcal{GP}(m^l(\cdot), k^l(\cdot, \cdot))$ (to simplify notation, the inter-layer noise is absorbed into the kernel $k^l(\cdot,\cdot)$ for $l \in \{1,\dots,L-1\}$). Inference in this model is possible through the use of the variational sparse approximation framework, where the ELBO is given by \citep{damianou_deep_2013,salimbeni_doubly_2017}
\begin{equation}
\mathcal{L}_\text{DGP} = \mathbb{E}_{q(\{\mathbf{F}^l, \mathbf{U}^l\}_{l=1}^L)}\left[\log \frac{p(\mathbf{Y}, \{\mathbf{F}^l, \mathbf{U}^l\}_{l=1}^L)}{q(\{\mathbf{F}^l, \mathbf{U}^l\}_{l=1}^L)}\right].
\label{eqn:dgp_elbo}
\end{equation}

\subsection{EVIDENCE LOWER BOUND (ELBO) DERIVATION}
\label{sec:detailed-elbo}
For the multitask DGP model defined in \eqref{eqn:expanded_joint} and the approximate posterior defined in \eqref{eqn:approx_posterior}, we are able to construct an ELBO using the formulation in \eqref{eqn:dgp_elbo}
\begin{align}
\mathcal{L} = &\mathbb{E}_{q(\{\mathbf{F}^t, \mathbf{\tilde{F}}^t, \{\mathbf{H}^t_j, \mathbf{\tilde{H}}^t_j\}_{j=1}^{J^t}\}_{t=1}^T, \{\mathbf{G}_i, \mathbf{\tilde{G}}_i\}_{i=1}^I)}\bigg[ \nonumber\\
&\log \frac{
    \prod_{t=1}^T[ \prod_{n=1}^{N^t}p(\mathbf{y}_n^t|\mathbf{F}^t)]
    \color{blue} \cancel{p(\mathbf{F}^t|\mathbf{\tilde{F}}^t, \Lambda^t)} p(\mathbf{\tilde{F}}^t)
    \color{green}\prod_{j=1}^{J^t} \cancel{p(\mathbf{H}_j^t|\mathbf{\tilde{H}}_j^t)} p(\mathbf{\tilde{H}}_j^t)\color{black}]
    \color{red}\prod_{i=1}^I [\cancel{p(\mathbf{G}_i|\mathbf{\tilde{G}}_i)} p(\mathbf{\tilde{G}}_i)]
}{
    \prod_{t=1}^T [\color{blue}\cancel{p(\mathbf{F}^t|\mathbf{\tilde{F}}^t, \Lambda^t)} q(\mathbf{\tilde{F}}^t)
    \color{green}\prod_{j=1}^{J^t} \cancel{p(\mathbf{H}_j^t|\mathbf{\tilde{H}}_j^t)} q(\mathbf{\tilde{H}}_j^t)\color{black}]
    \color{red}\prod_{i=1}^I [\cancel{p(\mathbf{G}_i|\mathbf{\tilde{G}}_i)} q(\mathbf{\tilde{G}}_i)]
}   \nonumber\\
\bigg].
\label{eqn:prem1-elbo}
\end{align}
Since the approximate posterior $q$ has a  factorised form \eqref{eqn:approx_posterior}, we can rewrite equation \eqref{eqn:prem1-elbo} as follow:
\begin{align}
\mathcal{L} =& \sum_{t=1}^T \sum_{n=1}^{N^t} \mathbb{E}_{q(\{\mathbf{F}^t, \mathbf{\tilde{F}}^t, \{\mathbf{H}^t_j, \mathbf{\tilde{H}}^t_j\}_{j=1}^{J^t}\}_{t=1}^T, \{\mathbf{G}_i, \mathbf{\tilde{G}}_i\}_{i=1}^I)}[\log p(\mathbf{y}_n^t | \mathbf{f}_n^t)] 
\color{blue} - \sum_{t=1}^T \text{KL}\left[q(\mathbf{\tilde{F}}^t) \middle\| p(\mathbf{\tilde{F}}^t)\right] \nonumber \\
&\color{green} - \sum_{t=1}^T \sum_{j=1}^{J^t} \text{KL}\left[q(\mathbf{\tilde{H}}^t_j) \middle\| p(\mathbf{\tilde{H}}_j^t)\right]
\color{red} -\sum_{i=1}^I \text{KL}\left[q(\mathbf{\tilde{G}}_i) \middle\| p(\mathbf{\tilde{G}}_i)\right]  \label{eqn:prem2-elbo}
\end{align}
The KL terms in equation \eqref{eqn:prem2-elbo} come from marginalising out the factors in the posterior that do not depend on the operand in the expectation terms. As discussed in \citet{salimbeni_doubly_2017}, for a DGP the $n$th marginal of the top layer only depends on the $n$th marginals of all the previous layers. Hence, in the case of the presented model we have
\begin{equation}
    q(\mathbf{f}_n^t) = \int q(\mathbf{f}_n^t | \boldsymbol{\lambda}_n^t) q(\boldsymbol{\lambda}_n^t) d\mathbf{c_n^t},
    \label{eqn:dgp-marginal}
\end{equation}
where $q(\mathbf{f}_n^t | \boldsymbol{\lambda}_n^t) = \int p(\mathbf{f}_n^t | \mathbf{\tilde{F}}^t; \boldsymbol{\lambda}_n^t, \mathbf{Z}_{\mathbf{F}^t}) q(\mathbf{\tilde{F}}^t)d{\mathbf{\tilde{F}}}^t$ and \\* $q(\boldsymbol{\lambda}_n^t) = \int \prod_{j=1}^{J^t} p(\mathbf{h}_{nj}^t|\mathbf{\tilde{H}}_j^t; \mathbf{x}_n^t, \mathbf{Z}_{\mathbf{H}_j^t}) q(\mathbf{\tilde{H}}_j^t) \prod_{i=1}^I p(\mathbf{g}_{ni}|\mathbf{\tilde{G}}_i; \mathbf{x}_n^t, \mathbf{Z}_{\mathbf{G}_i}) q(\mathbf{\tilde{G}}_i)d\mathbf{\tilde{H}}_j^t d\mathbf{\tilde{G}}_i$. By simplifying the expected log-likelihood term, we can rewrite \eqref{eqn:prem2-elbo} as follows (we reintroduced the dependence on the inducing locations $\mathbf{Z}$ back to the notation for completion)
\begin{align}
\mathcal{L} =& \sum_{t=1}^T \sum_{n=1}^{N^t} \mathbb{E}_{q(\mathbf{f}_n^t)}[\log p(\mathbf{y}_n^t | \mathbf{f}_n^t)] 
\color{blue} - \sum_{t=1}^T \text{KL}\left[q(\mathbf{\tilde{F}}^t) \middle\| p(\mathbf{\tilde{F}}^t; \mathbf{Z}_{\mathbf{F}^t})\right] \nonumber \\
&\color{green} - \sum_{t=1}^T \sum_{j=1}^{J^t} \text{KL}\left[q(\mathbf{\tilde{H}}^t_j) \middle\| p(\mathbf{\tilde{H}}_j^t; \mathbf{Z}_{\mathbf{H}_j^t})\right]
\color{red} -\sum_{i=1}^I \text{KL}\left[q(\mathbf{\tilde{G}}_i) \middle\| p(\mathbf{\tilde{G}}_i; \mathbf{Z}_{\mathbf{G}_i})\right] . \label{eqn:mtl-elbo2}
\end{align}

An important note is that the integral in \eqref{eqn:dgp-marginal} is intractable, hence we are unable to compute the expected log-likelihood term in \eqref{eqn:mtl-elbo2} in closed form. However, it is straightforward to sample from the marginal posterior $q(\mathbf{f}_n^t)$ using the re-parameterisation trick \citep{rezende_stochastic_2014,kingma_variational_2015}, allowing the approximate computation of the lower bound and its gradients by Monte Calro sampling.

%% file: sections/appendix_experiment_details.tex
\section{EXPERIMENT DETAILS}
\label{sec:experiment-details}
\subsection{MNIST VARIATIONS}
In this section, we detail the experimental setup used to obtain the results in \cref{sec:mnist}.

\paragraph{Data} We use the three background datasets from \url{https://sites.google.com/a/lisa.iro.umontreal.ca/public_static_twiki/variations-on-the-mnist-digits}. For each dataset. we sample 1000 images from the training split and combine them into a single 3 task multi-class classification dataset. For testing, we use the whole test split for each task.

\paragraph{Experimental procedure} We repeat the training dataset sub-sampling procedure 10 times to obtain 10 different training datasets. We fit the models on each of the of the training datasets and report the average accuracy and its standard error on the test split.

\paragraph{GP models initialisation} All GP models use the Mate\'rn-5/2 covariance function. For the covariance functions that act on the data, the kernel parameters are initialised to the same values for all the models. The lengthscale is set to $l=40$ and the signal variance $v=30$. One exception is the \mtldgp{} model where the kernel parameters in the shared component are initialised as before, but the task specific component is initialised to $l=20$ and $v=60$. This is done to break the symmetry with the shared component. Additionally, for deep model the kernel parameters on the top layer are initialised to $l=20$ and $v=30$. All models use the same initial set of inducing inputs per task, initialised as the k-means centroids from the training data. Each task admits 50 inducing inputs. Deep models have an inner layer of size 30.

\paragraph{Neural network architecture} All neural networks have inner layers with 128 neurons with ReLU activations. The standard networks use dropout with probability 0.2 and $L2$ regularisation. The Bayesian neural network use a standard normal prior on the weights. All weights are initialised with Glorot initialisation.

\paragraph{Optimisation} Shallow GP models use L-BFGS. Deep GP models use Adam with a learning rate 0.01 run for 10000 iterations. Mini-batching is used on the deep GP models with mini-batch size of 100. The standard neural networks use Adam with a learning rate of 0.001, 10000 iterations and mini-batch size of 64. The Bayesian neural networks use Adam with leaning rate 0.0001, 10000 iterations and mini-batch size of 64.

\subsection{SARCOS ROBOT INVERSE DYNAMICS}
In this section, we detail the experimental setup used to obtain the results in \cref{sec:sarcos}.

\paragraph{Data} We use the dataset from \url{http://www.gaussianprocess.org/gpml/data/}. This dataset relates to the inverse dynamics problem for a seven degrees-of-freedom anthropomorphic robot arm \citep{vijayakumar_statistical_2002,rasmussen_gaussian_2006}. The dataset consists of 44,484 training observations with 21 variables (7 joint positions, 7 joint velocities, 7 joint accelerations) and the corresponding real-valued torques for 7 joints, i.e. 7 tasks corresponding to each joint torque. Additionally, there are 4,449 testing examples with all the 7 joint torques available for each.

\paragraph{Experimental procedure} For $N$ in $\{100, 200, 500, 1000, 2000, 5000\}$, sample $N$ training data points from the training set. This constitutes the training set for one experimental run. For each of the $N$ sampled points select 1 of the 7 labels uniformly at random. Split the experimental training set into 7 according to which joint the label corresponds to. These 7 splits constitute 7 tasks. We train the models on the training set after standardising the features and the targets and test for all 7 labels on the test set.  For each $N$, we repeat the experimental procedure 10 times

\paragraph{GP models initialisation} All GP models use the Mate\'rn-5/2 covariance function. For the covariance functions that act on the data, the kernel parameters are initialised to the same values for all the models. The lengthscale is set to $l=10$ and the signal variance $v=1$. One exception is the \mtldgp{} model where the kernel parameters in the shared component are initialised as before, but the task specific component is initialised to $l=10$ and $v=0.5$. This is done to break the symmetry with the shared component. Additionally, for deep model the kernel parameters on the top layer are initialised to $l=10$ and $v=1$. All models use the same initial set of 100 inducing inputs distributed per task, initialised by random sampling from the training data. The likeihood noise variance is initialised to $\sigma^2 = 10^{-6}$. Deep models have an inner layer of size 10.

\paragraph{Neural network architecture} All neural networks have inner layers with 128 neurons with ReLU activations. The standard networks use dropout with probability 0.2 and $L2$ regularisation. The Bayesian neural network use a standard normal prior on the weights. All weights are initialised with Glorot initialisation.

\paragraph{Optimisation} Shallow GP models use L-BFGS for dataset sizes less than or equal to 1000, and Adam with the same settings as deep models otherwise. Deep GP models use Adam with a learning rate 0.01 run for 10000 iterations. Mini-batching is used on the deep GP models when the dataset size is greater than 1000 with mini-batch size of 500. The standard neural networks use Adam with a learning rate of 0.0001, 10000 iterations and mini-batch size of 64. The Bayesian neural networks use Adam with leaning rate 0.0001, 10000 iterations and mini-batch size of 64.

\subsection{FAIMS DIABETES DIAGNOSIS}
In this section, we detail the experimental setup used to obtain the results in \cref{sec:faims}.

\paragraph{Data} We use data from a case-control study of 125 patients who have been tested for diabetes. 48 out of 125 have been found to have diabetes (the disease group), while the rest are disease-free (the control group). The data consists of three experimental runs per patient, corresponding to sequential FAIMS analyses on the same urine sample. We treat each experimental run as a task, i.e. three binary classification tasks in total.

\paragraph{Experimental procedure} We perform Sparse Principal Component Analysis decomposition on the features selecting the first 20 principal components. We perform 10-fold cross-validation on 70:30 train-test splits.

\paragraph{GP models initialisation} All GP models use the Mate\'rn-5/2 covariance function. For the covariance functions that act on the data, the kernel parameters are initialised to the same values for all the models. The lengthscale is set to $l=0.5$ and the signal variance $v=1.5$. One exception is the \mtldgp{} model where the kernel parameters in the shared component are initialised as before, but the task specific component is initialised to $l=1.5$ and $v=1.5$. This is done to break the symmetry with the shared component. Additionally, for deep model the kernel parameters on the top layer are initialised to $l=1$ and $v=2$. All models use the same initial set of inducing inputs per task, initialised as the k-means centroids from the training data. Each task admits 30 inducing inputs. Deep models have an inner layer of size 3.

\paragraph{Neural network architecture} All neural networks have inner layers with 128 neurons with ReLU activations. The standard networks use dropout with probability 0.2 and $L2$ regularisation. The Bayesian neural network use a standard normal prior on the weights. All weights are initialised with Glorot initialisation.

\paragraph{Optimisation} Shallow GP models use L-BFGS. Deep GP models use Adam with a learning rate 0.001 run for 20000 iterations. The standard neural networks use Adam with a learning rate of 0.0001, 20000 iterations and mini-batch size of 64. The Bayesian neural networks use Adam with leaning rate 0.0001, 20000 iterations and mini-batch size of 64.

%% file: sections/appendix_mnist.tex
\section{FURTHER RESULTS}
\subsection{MNIST VARIATIONS}
\begin{table}[h!]
\caption{Classification accuracy on the MNIST variations experiment using all three tasks. Higher is better}
\vspace{.3in}
\adjustbox{max width=\textwidth}{%
\centering
\begin{tabular}{lcccccccccc} 
\toprule
\multicolumn{11}{c}{Accuracy} \\
\cmidrule(r){2-11}
 Task & \sharedmtldgp & \mtldgp & \icmdgp & \stldgp & \icmgp & \stlgp & \mtlannTwo & \mtlannThree & \mtlbnnTwo & \mtlbnnThree \\
\midrule
Standard MNIST & $\mathbf{0.89(0.0})$ & $0.87(0.00)$ & $0.80(0.08)$ & $0.85(0.00)$ & $0.87(0.00)$ & $0.87(0.00)$ & $0.88(0.00)$ & $0.88(0.00)$ & $0.83(0.00)$ & $0.83(0.00)$  \\
Random BG MNIST & $\mathbf{0.64(0.0})$ & $0.61(0.01)$ & $0.61(0.03)$ & $0.18(0.00)$ & $0.59(0.04)$ & $0.19(0.00)$ & $0.59(0.00)$ & $0.59(0.00)$ & $0.51(0.01)$ & $0.49(0.01)$  \\
Images BG MNIST & $\mathbf{0.78(0.0})$ & $0.74(0.00)$ & $0.70(0.07)$ & $0.19(0.01)$ & $0.58(0.06)$ & $0.21(0.01)$ & $0.62(0.00)$ & $0.61(0.00)$ & $0.53(0.01)$ & $0.53(0.01)$  \\
All & $\mathbf{0.77(0.0})$ & $0.74(0.00)$ & $0.70(0.06)$ & $0.41(0.01)$ & $0.68(0.04)$ & $0.42(0.00)$ & $0.70(0.00)$ & $0.69(0.00)$ & $0.63(0.01)$ & $0.61(0.01)$  \\
\bottomrule
\end{tabular}}
\vspace{.3in}
\end{table}

%% file: sections/appendix_sarcos.tex
\subsection{SARCOS ROBOT INVERSE DYNAMICS}
\label{sec:appendix_sarcos}
\begin{table*}[ht!]
\small
\caption{Average RMSE scores on the SARCOS score over 7 tasks. The figures presented are the mean score (and standard error) over 10 runs. The lowest statistically significant scores based on a Wilcoxon test are presented in boldface. Lower is better.}
\vspace{.3in}
\centering
\begin{tabular}{lccccccc}  
\toprule
& \multicolumn{6}{c}{Number of Training Inputs} \\
\cmidrule(r){2-7}
 & 100 & 200 & 500 & 1000 & 2000 & 5000\\
\midrule
\sharedmtldgp & $0.78(0.05)$ & $0.64(0.05)$ & $0.53(0.03)$ & $\mathbf{0.31(0.03)}$ & $\mathbf{0.26(0.02)}$ & $\mathbf{0.22(0.02)}$  \\
\mtldgp & $0.80(0.05)$ & $0.69(0.05)$ & $0.51(0.05)$ & $\mathbf{0.31(0.03)}$ & $\mathbf{0.26(0.03)}$ & $0.23(0.02)$  \\
\icmdgp & $0.78(0.07)$ & $0.64(0.05)$ & $0.52(0.03)$ & $0.33(0.03)$ & $0.27(0.03)$ & $0.23(0.02)$  \\
\stldgp & $0.78(0.06)$ & $0.64(0.05)$ & $0.53(0.03)$ & $0.47(0.02)$ & $0.36(0.03)$ & $0.26(0.02)$  \\
\icmgp & $0.98(0.01)$ & $0.96(0.01)$ & $0.93(0.03)$ & $0.86(0.06)$ & $0.29(0.03)$ & $0.26(0.03)$  \\
\stlgp & $0.98(0.02)$ & $0.95(0.02)$ & $0.92(0.02)$ & $0.89(0.05)$ & $0.80(0.07)$ & $0.63(0.07)$  \\
\mtlannTwo & $\mathbf{0.73(0.06)}$ & $\mathbf{0.58(0.04)}$ & $\mathbf{0.41(0.03)}$ & $0.34(0.02)$ & $0.31(0.02)$ & $0.30(0.02)$  \\
\mtlannThree & $0.77(0.07)$ & $0.62(0.05)$ & $0.45(0.03)$ & $0.39(0.02)$ & $0.36(0.02)$ & $0.35(0.02)$  \\
\mtlbnnTwo & $0.80(0.07)$ & $0.62(0.05)$ & $0.44(0.03)$ & $0.34(0.02)$ & $0.27(0.02)$ & $\mathbf{0.22(0.02)}$  \\
\mtlbnnThree & $0.79(0.07)$ & $0.64(0.05)$ & $0.46(0.03)$ & $0.36(0.02)$ & $0.29(0.02)$ & $0.24(0.02)$  \\
\bottomrule
\end{tabular}
\label{table:rmse}
\vspace{.3in}
\end{table*}

\begin{table}
\caption{ELBO computation cost in CPU time for different multitask GP models. These are timed on a system with the folowing specifications: OS: Linux 4.9.202-1-Manjaro with X96-64 instruction set. CPU: Intel Core i7-4712HQ @ 2.30GHz. RAM: 16GB. }
\label{table:timing}
\vspace{.3in}
\centering
\begin{tabular}{lc}  
\toprule
\multicolumn{2}{c}{SARCOS: ELBO computation} \\
 & time (batch size = 500) \\
\midrule
 Model & Wall-clock time \\
 & in milliseconds \\
\midrule
\sharedmtldgp & $7.89(0.15)$ \\
\mtldgp & $8.89(0.25)$ \\
\icmdgp & $7.67(0.29)$ \\
\stldgp & $6.43(0.16)$ \\
\icmgp & $2.41(0.04)$ \\
\stlgp & $3.22(0.06)$ \\
\bottomrule
\end{tabular}
\end{table}

\begin{figure}[ht!]
\centering
\includegraphics[width=0.6\linewidth]{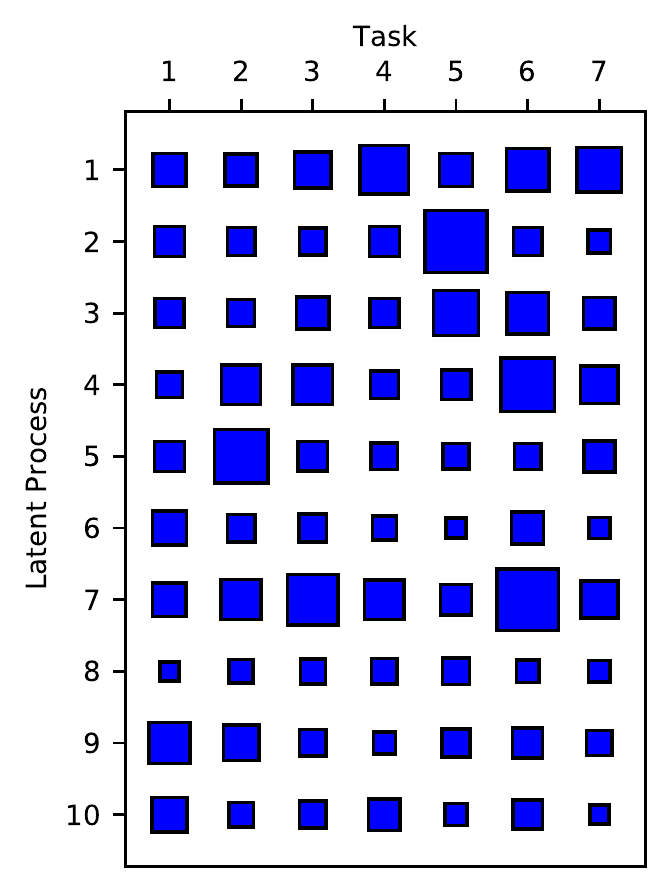}
\caption{Hinton diagram illustrating the ARD weights for on of the sMDGP runs on Sarcos with 5000 training datapoints. Blocks in blue are shared latent processes, while blocks in red are task specific. This indicates that the model assigns different weights for the latent processes for each task.}
\label{fig:hinton2}
\end{figure}

%% file: main.bbl
\begin{thebibliography}{41}
\providecommand{\natexlab}[1]{#1}
\providecommand{\url}[1]{\texttt{#1}}
\expandafter\ifx\csname urlstyle\endcsname\relax
  \providecommand{\doi}[1]{doi: #1}\else
  \providecommand{\doi}{doi: \begingroup \urlstyle{rm}\Url}\fi

\bibitem[Aglietti et~al.(2019)Aglietti, Damoulas, and Bonilla]{aglietti19a}
V.~Aglietti, T.~Damoulas, and E.~V. Bonilla.
\newblock Efficient inference in multi-task cox process models.
\newblock In \emph{Proceedings of Machine Learning Research}, 2019.

\bibitem[Alaa and van~der Schaar(2017)]{alaa_deep_2017}
A.~M. Alaa and M.~van~der Schaar.
\newblock Deep {Multi}-task {Gaussian} {Processes} for {Survival} {Analysis}
  with {Competing} {Risks}.
\newblock In \emph{Advances in {Neural} {Information} {Processing} {Systems}},
  Long Beach, California, 2017.

\bibitem[Alvarez and Lawrence(2009)]{alvarez_sparse_2009}
M.~Alvarez and N.~D. Lawrence.
\newblock Sparse convolved {Gaussian} {Process} for multi-output regression.
\newblock In \emph{Advances in {Neural} {Information} {Processing} {Systems}},
  Vancouver, Canada, 2009.

\bibitem[Alvarez et~al.(2011)Alvarez, Rosasco, and
  Lawrence]{alvarez_kernels_2011}
M.~Alvarez, L.~Rosasco, and N.~D. Lawrence.
\newblock Kernels for {Vector}-{Valued} {Functions}: a {Review}.
\newblock Technical {Report}, MIT, 2011.

\bibitem[Argyriou et~al.(2007)Argyriou, Evgeniou, and
  Pontil]{argyriou2007multi}
A.~Argyriou, T.~Evgeniou, and M.~Pontil.
\newblock Multi-task feature learning.
\newblock In \emph{Advances in neural information processing systems}, 2007.

\bibitem[Bakker and Heskes(2003)]{bakker_task_2003}
B.~Bakker and T.~Heskes.
\newblock Task clustering and gating for {Bayesian} {Multitask} {Learning}.
\newblock \emph{Journal of Machine Learning Research}, 2003.

\bibitem[Bonilla et~al.(2007)Bonilla, Chai, and
  Williams]{bonilla_multi-task_2007}
E.~V. Bonilla, K.~M. Chai, and C.~Williams.
\newblock Multi-task {Gaussian} process prediction.
\newblock In \emph{Advances in {Neural} {Information} {Processing} {Systems}},
  pages 153--160, Vancouver, Canada, 2007.

\bibitem[Boyle and Frean(2004)]{boyle_dependent_2004}
P.~Boyle and M.~Frean.
\newblock Dependent {Gaussian} {Processes}.
\newblock In \emph{Advances in {Neural} {Information} {Processing} {Systems}},
  Vancouver, Canada, 2004.

\bibitem[Calder and Cressie(2007)]{calder2007some}
C.~A. Calder and N.~Cressie.
\newblock Some topics in convolution-based spatial modeling.
\newblock \emph{Proceedings of the 56th Session of the International Statistics
  Institute}, 2007.

\bibitem[Caruana(1997)]{caruana_multitask_1997}
R.~Caruana.
\newblock Multitask {Learning}.
\newblock \emph{Machine Learning}, 28\penalty0 (1):\penalty0 41--75, 1997.

\bibitem[Covington et~al.(2015)Covington, Schee, Edge, Boyle, Savage, and
  Arasaradnam]{covington_application_2015}
J.~A. Covington, M.~P. v.~d. Schee, A.~S.~L. Edge, B.~Boyle, R.~S. Savage, and
  R.~P. Arasaradnam.
\newblock The application of {FAIMS} gas analysis in medical diagnostics.
\newblock \emph{Analyst}, 2015.

\bibitem[Damianou and Lawrence(2013)]{damianou_deep_2013}
A.~C. Damianou and N.~D. Lawrence.
\newblock Deep {Gaussian} {Processes}.
\newblock In \emph{Proceedings of 16th {International} {Conference} on
  {Artificial} {Intelligence} and {Statistics} ({AISTATS})}, Scottsdale, USA,
  2013. JMLR.

\bibitem[Damianou et~al.(2012)Damianou, Ek, Titsias, and
  Lawrence]{damianou_manifold_2012}
A.~C. Damianou, C.~H. Ek, M.~K. Titsias, and N.~D. Lawrence.
\newblock Manifold {Relevance} {Determination}.
\newblock In \emph{Proceedings of the 29th {International} {Conference} on
  {Machine} {Learning}}, Edinburgh, UK, 2012.

\bibitem[Evgeniou and Pontil(2004)]{evgeniou2004regularized}
T.~Evgeniou and M.~Pontil.
\newblock Regularized multi--task learning.
\newblock In \emph{Proceedings of the tenth ACM SIGKDD international conference
  on Knowledge discovery and data mining}, 2004.

\bibitem[Gal and Ghahramani(2016)]{gal2016dropout}
Y.~Gal and Z.~Ghahramani.
\newblock Dropout as a bayesian approximation: Representing model uncertainty
  in deep learning.
\newblock In \emph{international conference on machine learning}, 2016.

\bibitem[Goovaerts(1997)]{goovaerts_geostatistics_1997}
P.~Goovaerts.
\newblock \emph{Geostatistics for {Natural} {Resources} {Evaluation}}.
\newblock Oxford University Press, 1997.
\newblock ISBN 978-0-19-511538-3.

\bibitem[Hastie et~al.(2009)Hastie, Tibshirani, and
  Friedman]{hastie_elements_2009}
T.~Hastie, R.~Tibshirani, and J.~H. Friedman.
\newblock \emph{The {Elements} of {Statistical} {Learning}: {Data} {Mining},
  {Inference}, and {Prediction}}.
\newblock Springer, 2009.
\newblock ISBN 978-0-387-84884-6.

\bibitem[Hensman et~al.(2013)Hensman, Fusi, and
  Lawrence]{hensman_gaussian_2013}
J.~Hensman, N.~Fusi, and N.~D. Lawrence.
\newblock Gaussian processes for big data.
\newblock In \emph{Proceedings of the {Conference} on {Uncertainty} in
  {Artificial} {Intelligence}}, Bellevue, Washington, USA, 2013.

\bibitem[Higdon(1998)]{higdon1998process}
D.~Higdon.
\newblock A process-convolution approach to modelling temperatures in the north
  atlantic ocean.
\newblock \emph{Environmental and Ecological Statistics}, 1998.

\bibitem[Higdon et~al.(1999)Higdon, Swall, and Kern]{higdon1999non}
D.~Higdon, J.~Swall, and J.~Kern.
\newblock Non-stationary spatial modeling.
\newblock \emph{Bayesian statistics}, 1999.

\bibitem[Kandemir(2015)]{kandemir_asymmetric_2015}
M.~Kandemir.
\newblock Asymmetric {Transfer} {Learning} with {Deep} {Gaussian} {Processes}.
\newblock In \emph{Proceedings of the 32nd {International} {Conference} on
  {Machine} {Learning}}, Lille, France, 2015.

\bibitem[Kingma and Ba(2015)]{kingma_adam:_2015}
D.~P. Kingma and J.~Ba.
\newblock Adam: {A} method for stochastic optimization.
\newblock In \emph{International {Conference} on {Learning} {Representations}},
  San Diego, California, 2015.

\bibitem[Kingma et~al.(2015)Kingma, Salimans, and
  Welling]{kingma_variational_2015}
D.~P. Kingma, T.~Salimans, and M.~Welling.
\newblock Variational {Dropout} and the {Local} {Reparameterization} {Trick}.
\newblock In \emph{Advances in {Neural} {Information} {Processing} {Systems}},
  2015.

\bibitem[Knoblauch et~al.(2019)Knoblauch, Jewson, and
  Damoulas]{knoblauch2019generalized}
J.~Knoblauch, J.~Jewson, and T.~Damoulas.
\newblock Generalized variational inference.
\newblock \emph{arXiv preprint arXiv:1904.02063}, 2019.

\bibitem[Lecun et~al.(1998)Lecun, Bottou, Bengio, and
  Haffner]{lecun_gradient-based_1998}
Y.~Lecun, L.~Bottou, Y.~Bengio, and P.~Haffner.
\newblock Gradient-based learning applied to document recognition.
\newblock \emph{Proceedings of the IEEE}, 1998.

\bibitem[Mairal et~al.(2009)Mairal, Bach, Ponce, and
  Sapiro]{mairal_online_2009}
J.~Mairal, F.~Bach, J.~Ponce, and G.~Sapiro.
\newblock Online dictionary learning for sparse coding.
\newblock In \emph{Proceedings of the 26th {International} {Conference} on
  {Machine} {Learning}}, Montreal, Canada, 2009. ACM.

\bibitem[Martinez-Vernon et~al.(2018)Martinez-Vernon, Covington, Arasaradnam,
  Esfahani, O’Connell, Kyrou, and Savage]{martinez-vernon_improved_2018}
A.~S. Martinez-Vernon, J.~A. Covington, R.~P. Arasaradnam, S.~Esfahani,
  N.~O’Connell, I.~Kyrou, and R.~S. Savage.
\newblock An improved machine learning pipeline for urinary volatiles disease
  detection: {Diagnosing} diabetes.
\newblock \emph{PLOS ONE}, 2018.

\bibitem[Matthews et~al.(2017)Matthews, Alexander, Van Der~Wilk, Nickson,
  Fujii, Boukouvalas, León-Villagrá, Ghahramani, and
  Hensman]{matthews_gpflow:_2017}
D.~G. Matthews, G.~Alexander, M.~Van Der~Wilk, T.~Nickson, K.~Fujii,
  A.~Boukouvalas, P.~León-Villagrá, Z.~Ghahramani, and J.~Hensman.
\newblock {GPflow}: {A} {Gaussian} process library using {TensorFlow}.
\newblock \emph{The Journal of Machine Learning Research}, 2017.

\bibitem[Nguyen and Bonilla(2013)]{nguyen_efficient_2013}
T.~Nguyen and E.~Bonilla.
\newblock Efficient {Variational} {Inference} for {Gaussian} {Process}
  {Regression} {Networks}.
\newblock In \emph{Proceedings of 16th {International} {Conference} on
  {Artificial} {Intelligence} and {Statistics} ({AISTATS})}, 2013.

\bibitem[Nguyen and Bonilla(2014)]{nguyen_collaborative_2014}
T.~V. Nguyen and E.~V. Bonilla.
\newblock Collaborative {Multi}-output {Gaussian} {Processes}.
\newblock In \emph{Proceedings of the {Conference} on {Uncertainty} in
  {Artificial} {Intelligence}}, Quebec City, Canada, 2014.

\bibitem[Nocedal and Wright(2000)]{nocedal_numerical_2000}
J.~Nocedal and S.~J. Wright.
\newblock \emph{Numerical optimization}.
\newblock Springer series in operations research. Springer, New York, NY, 2000.
\newblock ISBN 978-0-387-98793-4.

\bibitem[Rasmussen and Williams(2006)]{rasmussen_gaussian_2006}
C.~E. Rasmussen and C.~K.~I. Williams.
\newblock \emph{Gaussian processes for machine learning}.
\newblock Adaptive computation and machine learning. MIT Press, Cambridge,
  Mass, 2006.

\bibitem[Requeima et~al.(2019)Requeima, Tebbutt, Bruinsma, and
  Turner]{requeima2019gaussian}
J.~Requeima, W.~Tebbutt, W.~Bruinsma, and R.~E. Turner.
\newblock The gaussian process autoregressive regression model (gpar).
\newblock In \emph{The 22nd International Conference on Artificial Intelligence
  and Statistics}, 2019.

\bibitem[Rezende et~al.(2014)Rezende, Mohamed, and
  Wierstra]{rezende_stochastic_2014}
D.~J. Rezende, S.~Mohamed, and D.~Wierstra.
\newblock Stochastic {Backpropagation} and {Approximate} {Inference} in {Deep}
  {Generative} {Models}.
\newblock In \emph{Proceedings of the 31st {International} {Conference} on
  {Machine} {Learning}}, 2014.

\bibitem[Salimbeni and Deisenroth(2017)]{salimbeni_doubly_2017}
H.~Salimbeni and M.~Deisenroth.
\newblock Doubly stochastic variational inference for deep gaussian processes.
\newblock In \emph{Advances in {Neural} {Information} {Processing} {Systems}},
  Long Beach, California, 2017.

\bibitem[Skolidis and Sanguinetti(2011)]{skolidis_bayesian_2011}
G.~Skolidis and G.~Sanguinetti.
\newblock Bayesian {Multitask} {Classification} {With} {Gaussian} {Process}
  {Priors}.
\newblock \emph{IEEE Transactions on Neural Networks}, 22\penalty0
  (12):\penalty0 2011--2021, 2011.

\bibitem[Teh et~al.(2005)Teh, Seeger, and Jordan]{whye_teh_semiparametric_2005}
Y.~W. Teh, M.~Seeger, and M.~Jordan, I.
\newblock Semiparametric {Latent} {Factor} {Models}.
\newblock In \emph{Proceedings of 10th {International} {Conference} on
  {Artificial} {Intelligence} and {Statistics} ({AISTATS})}, 2005.

\bibitem[Titsias(2009)]{titsias_variational_2009}
M.~K. Titsias.
\newblock Variational {Learning} of {Inducing} {Variables} in {Sparse}
  {Gaussian} {Processes}.
\newblock In \emph{Proceedings of 12th {International} {Conference} on
  {Artificial} {Intelligence} and {Statistics} ({AISTATS})}, volume~5,
  Clearwater Beach, Florida, USA, 2009.

\bibitem[Titsias and Lazaro-Gredilla(2011)]{titsias_spike_2011}
M.~K. Titsias and M.~Lazaro-Gredilla.
\newblock Spike and slab variational inference for multi-task and multiple
  kernel learning.
\newblock In \emph{Advances in {Neural} {Information} {Processing} {Systems}},
  Granada, Spain, 2011.

\bibitem[Vijayakumar et~al.(2002)Vijayakumar, D'souza, Shibata, Conradt, and
  Schaal]{vijayakumar_statistical_2002}
S.~Vijayakumar, A.~D'souza, T.~Shibata, J.~Conradt, and S.~Schaal.
\newblock Statistical learning for humanoid robots.
\newblock \emph{Autonomous Robots}, 2002.

\bibitem[Wilson et~al.(2012)Wilson, Knowles, and
  Ghahramani]{wilson_gaussian_2012}
A.~G. Wilson, D.~A. Knowles, and Z.~Ghahramani.
\newblock Gaussian {Process} {Regresssion} {Networks}.
\newblock In \emph{Proceedings of the 29th {International} {Conference} on
  {Machine} {Learning}}, Edinburgh, UK, 2012.

\end{thebibliography}
